\documentclass{article}
\usepackage{jfrExamplee}
\usepackage{graphicx}
\usepackage{subfig}
\usepackage{apalike}
\usepackage{setspace}
\usepackage{amsmath}
\usepackage{url}
\title{Singularity-Free Guiding Vector Field over Bézier's Curves applied to Rovers Path Planning and Path following.   }

\author{Alfredo González-Calvin%\thanks{ Use footnote for providing further information
\\
Dept. Arquitectura de Computadores y Automática\\
Facultad de Ciencias Físicas. Universidad Complutense\\
Madrid, Spain  \\
\texttt{alfredgo@ucm.es} \\
\And
Lía García-Pérez \\
Dept. Arquitectura de Computadores y Automática\\
Facultad de Ciencias Físicas. Universidad Complutense\\
Madrid, Spain  \\
\texttt{liagar05@ucm.es} \\
\AND
Juan F. Jiménez\\
Dept. Arquitectura de Computadores y Automática\\
Facultad de Ciencias Físicas. Universidad Complutense\\
Madrid, Spain  \\
\texttt{juan.jimenez@fis.ucm.es} \\
}
\usepackage{comment}
\usepackage{amsmath,amssymb,amsfonts}
\usepackage{amsthm}
\usepackage{algorithmic}
\usepackage{textcomp}
\usepackage{xcolor}
\usepackage{tikz}
\usetikzlibrary{calc}
\usetikzlibrary{shapes.geometric}
\usetikzlibrary{shapes.multipart}\def\BibTeX{{\rm B\kern-.05em{\sc i\kern-.025em b}\kern-.08em
    T\kern-.1667em\lower.7ex\hbox{E}\kern-.125emX}}

\newtheorem{proposition}{Proposition}
\newtheorem{problem}{Problem}
\begin{document}

\maketitle

\begin{abstract}
This paper presents a guidance algorithm for solving the problem of following parametric paths, as well as a curvature-varying speed setpoint for land-based car-type wheeled mobile robots (WMRs). The guidance algorithm relies on Singularity-Free Guiding Vector Fields SF-GVF. This novel GVF approach expands the desired robot path and the Guiding vector field to a higher dimensional space, in which an angular control function can be found to ensure global asymptotic convergence to the desired parametric path while avoiding field singularities. In SF-GVF, paths should follow a parametric definition. This feature makes using Bezier's curves attractive to define the robot's desired path. The curvature-varying speed setpoint, combined with the guidance algorithm, eases the convergence to the path when physical restrictions exist, such as minimal turning radius or maximal lateral acceleration. We provide theoretical results, simulations, and outdoor experiments using a WMR platform assembled with off-the-shelf components. The small Rover (WMR) selected provides an easy-to-use non-holonomic platform for the experiments. The results could be extrapolated to full-scale or more complex vehicles, providing the necessary vehicle control system adaptations, while the GVF algorithm would remain the same. 

\end{abstract}

\section{Introduction}\label{sec:intro}

Autonomous Mobile Robots (AMRs) have become a broad research area with applications in many fields; from automatic storage systems in warehouses, environment monitoring, post-catastrophe inspection, and emergency rescue to planetary exploration, the presence of AMRs is ubiquitous. A review of the categories and applications of AMRs can be found in \cite{rubiovalero}.

Following a classical scheme, \cite[page 10]{introamv}, the mobility control of the AMR could be divided into two main tasks: perception and motion control. We are interested in motion control since it makes the AMR follow a planned path(s) (path following).

There are certain types of AMR applications in which the robots are required to pass through specified points that an expert user defines. For example, environmental monitoring robots \cite{hitz2017adaptive} that take samples or measurements at points of interest, surveillance robots that need to visit specific locations \cite{thakur2013}, agricultural robots that must apply treatments at designated locations \cite{bechar2016agricultural,bechar2017agricultural}, or delivery robots \cite{Neil2015} that have to deliver products to defined locations. Usually, the robot's human operators have prior knowledge of the mission. They can plan the robot's path by breaking it into feasible trajectories. This can be done using maps or photographs of the terrain the robots will traverse and defining a set of waypoints that the Rover should visit or at least pass close to. The robot, in turn, must generate and follow a trajectory that ensures these points are reached.
In some cases, it is helpful for the user to define, in addition to the waypoints, an area within which the trajectory is restricted, for example, for safety reasons. Think, for instance, about rovers for outdoor exploration or patrol. They should be able to cope with unstructured environments where slopes, terrain irregularities, and obstacles could hamper the Rover's performance. 

However, for a robot to be able to follow a path visiting specific waypoints autonomously, it is not enough to define the trajectory between them. To make it an autonomous system, an AMR must be able to drive itself, be aware of its state, avoid obstacles, etc., and ultimately reach its predetermined goals. For this purpose, once a suitable trajectory has been defined for the AMR, it is essential to provide a motion control algorithm that allows it to follow the trajectory correctly. The control algorithm can be divided into Guidance and Navigation.

The guidance controller provides the desired setpoints (e.g., position, heading and speed) to take the robot from its current state (i.e., position, attitude and speed) to the desired trajectory. When planning a trajectory, we may consider the mission the robot has to accomplish as the starting point. This planning process must take into account the kinematic and dynamic constraints. Our robot is a typical nonholonomic system, a (four) Wheeled Mobile Robot, WMR, with a minimum turning radius and maximum acceleration limits to avoid wheel slippage, lateral skidding or robot overturning. The latter two could be estimated using an Inertial Unit on board \cite{nour2023}. 

In some cases, the upper and lower speed limits are mission-dependent, while in others, the speed limit is fixed during the whole mission. Using the first approach, we can facilitate the operator's search for feasible trajectories by linking the speed control to the curvature of the trajectory, thus connecting it to the guidance controllers. Therefore, it is sufficient to define waypoints to design a reasonable path, avoiding significant obstacles and sharp turns, and leave it to the guidance system to adjust the speed to follow the resulting trajectory.

The AMR navigation comprises headings and speed control to allow the robot to follow the desired trajectory. It is linked to the mechanical characteristics of the robot and the specific onboard actuators and sensors.

This paper outlines the development, simulation, and field experimentation of a navigation system based on waypoints. This system is implemented and tested on a small, land-based, four-wheeled mobile robot (WMR) with a low-cost onboard electronic system. We give this particular type of WMR the generic name of \textit{Rover} because it can move over rough terrain.

The Rover has to pass through the waypoints determined by the user. These waypoints are connected by Bézier curves, whose first and last points are the waypoints, while the rest of them (control points) allow the route to remain within safe navigation zones.
The Rover's task is to follow the trajectory calculated using the Bézier curves. The user can modify waypoints and control points during the mission and send them to the Rover. When the Rover receives them, a new trajectory is generated. The path-following system, consisting of guidance and speed control, computes the appropriate commands to enable the Rover to follow the trajectory through the waypoints. We use Guidance Vector Fields (GVF), in particular, the novel Singular Free Parametric GVF (SF-GVF) methodology \cite{weijiath}, to set the heading and the curvature of the trajectory to set the speed. A non-linear control strategy for the heading ensures mathematical convergence to the path. In addition, we use a linear controller for the speed.

As we shall describe later in the paper, SF-GVF requires defining parametric curves to determine the robot's path. However, this can be challenging in real-world applications, especially when the robot moves in non-structured environments.
To address this issue, we propose using Bezier curves, which are parametric curves well-suited for defining the robot's path in SF-GVF. This approach can make it easier to apply SF-GVF to real-world scenarios. The user only needs to specify the control points for the Bezier curve and then check whether the resulting path meets their requirements.

Results of the guidance and navigation controllers are presented first in simulation and after with a real Rover in a natural environment. In both cases, the Paparazzi environment was used \cite{gati2013open}. Paparazzi autopilot has been used together with GVF for UAVs \cite{KAPITANYUK2017},\cite{deMarina2017}.

In the following sections of the paper, we will progressively develop our proposal according to the following sequence: Section \ref{sec:model} describes a simple Rover model that will be employed later to build the guidance control system. Section \ref{sec:GVF_Rover} deals with the SF-GVF guidance approach, the Rover steering control description and its stability properties. Section \ref{sec:PathPlanning} explains the use of Bézier curves in path planning and discusses their main advantages and disadvantages. Section \ref{sec:speed_control} presents the speed controller used in our Rover and the curvature-varying setpoint. Section \ref{sec:results} is devoted to simulated and experimental results, briefly introducing the Paparazzi ecosystem, an open-source software platform we use to develop Rover onboard code and simulate and monitor experiments with a team of small Rovers. Eventually, we close the work in section \ref{sec:conclusions}, drawing some conclusions.

 \section{Rover model}\label{sec:model}
Our Rover is a four-wheeled WMR with a spring suspension system to ensure maximum contact between the wheels and the ground. Two fixed rear wheels on a joint axle and two front steering wheels, a car-like or type (1,1) according to \cite{siciliano2008springer} robot classification.

\subsection{Basic model assumptions}
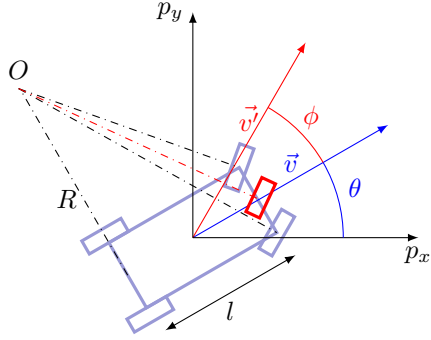
\begin{figure}
\centering
\begin{tikzpicture}
\draw(0,0)[rotate around={30:(0,0)},very thick,draw = blue!60!black!40]rectangle(2,1);
\draw(-0.3,-0.2)[rotate around={30:(0,0)},very thick,draw = blue!60!black!40]rectangle(0.3,0.0);
\draw(-0.3,1)[rotate around={30:(0,0)},very thick,draw = blue!60!black!40]rectangle(0.3,1.2);
\draw({2*cos(30)-0.3},{2*sin(30)-0.1})[rotate around={60:({2*cos(30)},{2*sin(30)})},very thick,draw = blue!60!black!40]rectangle({2*cos(30)+0.3},{2*sin(30)+0.1});
\draw({2*cos(30)-sin(30)-0.3},{2*sin(30)+cos(30)-0.1})[rotate around={70:({2*cos(30)-sin(30)},{2*sin(30)+cos(30)})},very thick,draw = blue!60!black!40]rectangle({2*cos(30)-sin(30)+0.3},{2*sin(30)+cos(30)+0.1});
\draw[blue]({cos(30)-0.5*sin(30)},{sin(30)+0.5*cos(30)})--node[above]{$\vec{v}$}({4*cos(30)-0.5*sin(30)},{4*sin(30)+0.5*cos(30)})[-latex];
\draw[red]({cos(30)-0.5*sin(30)},{sin(30)+0.5*cos(30)})--node[above]{$\vec{v'}$}({cos(30)-0.5*sin(30)+3*cos(60},{sin(30)+0.5*cos(30)+3*sin(60})[-latex];

\draw({cos(30)-0.5*sin(30)},{sin(30)+0.5*cos(30)})--({3+cos(30)-0.5*sin(30)},{sin(30)+0.5*cos(30)})[-latex]node[anchor=north]{$p_x$};
\draw({cos(30)-0.5*sin(30)},{sin(30)+0.5*cos(30)})--({cos(30)-0.5*sin(30)},{3+sin(30)+0.5*cos(30)})[-latex]node[anchor=east]{$p_y$};

\draw[blue]({2+cos(30)-0.5*sin(30)},{sin(30)+0.5*cos(30)})arc(0:30:2);
\draw[red]({3*cos(30)-0.5*sin(30)},{3*sin(30)+0.5*cos(30)})arc(30:60:2);
\draw[blue]({3.2*cos(30)},{3.2*sin(30)}) node[]{$\theta$};
\draw[red]({3.4*cos(50)},{3.3*sin(50)}) node[]{$\phi$};
\draw[latex-latex](0.25,-0.3)--node[below]{$l$}({2*cos(30)+0.25},{2*sin(30)-0.3});

\draw({2*cos(30)-0.55},{2*sin(30)+0.5})[rotate around={65:({2*cos(30)-0.15},{2*sin(30)+0.6})},very thick,draw = red]rectangle({2*cos(30)-0.05},{2*sin(30)+0.7});

\draw[dashdotted]({0.5*cos(120)},{0.5*sin(120})--({3.35*cos(120)},{3.35*sin(120});
\draw({2*cos(122)},{2*sin(122})[anchor=north]node{$R$};
\draw[dashdotted]({2.5*cos(120)-0.45},{2*sin(30)+cos(30)+1.05})--({2*cos(30)-sin(30)},{2*sin(30)+cos(30)});
\draw[dashdotted]({2.5*cos(120)-0.45},{2*sin(30)+cos(30)+1.05})node[anchor=south]{$O$}--({2*cos(30)},{2*sin(30)});
\draw[dashdotted,draw = red]({2.5*cos(120)-0.45},{2*sin(30)+cos(30)+1.05})--({2*cos(30)-0.25},{2*sin(30)+0.45});

\end{tikzpicture}
\caption{Representation of a simple four-wheeled car with wheelbase $l$. The car is shown in light blue, while the virtual front wheel is shown in red. The direction of motion is shown with a blue line, and the direction of the virtual wheel is shown in red. }
\label{fig:vehir}
\end{figure}

Figure \ref{fig:vehir} depicted an outline of a four-wheeled vehicle. As it is known, a four-wheeled car will not skid if the longitudinal axis of each wheel is aligned with the circumferences they trace around to the center of rotation. For this purpose, an Ackermann steering linkage can be employed, in which each wheel turns a different angle when tracing a curve, according to well-known equations \cite{siciliano2008springer}. However, when conditions are ideal (i.e., no skid), the two wheels of an Ackerman drive can be replaced by a virtual single wheel situated at the center of the front axis, that is, in the direction of motion of the Rover. 
% Rover vehicle
% Rover Vehicle
Therefore, in figure \ref{fig:vehir}, the angle $\theta$ represents the Rover's heading in an inertial frame $W(p), p = (p_x,p_y) \in \mathbb{R}^2$ attached to the earth and the angle $\phi$ is the angle between the longitudinal axis of the car (i.e., its direction of motion) and the virtual single wheel (shown in red). Then, the velocity can be computed as:
\begin{equation}
    \begin{cases}
        \dot p_x &= v\cos(\theta) \\
        \dot p_y &= v\sin(\theta)
    \end{cases}
\end{equation}
where $v$ is the speed in the direction of motion.
Now let $r$ be the distance travelled by the car (by its rear axis), and let $R$ be the instantaneous radius of rotation. An infinitesimal displacement in the distance implies $dr = R d\theta$, and it can be seen that $R = \frac{l}{\tan(\phi)}$. Then $\dot r = v = \frac{l}{\tan(\phi)}\dot \theta$, where $l$ represents the wheelbase of the vehicle. Finally, assume that the control input is the angular rotation of the virtual front wheel $\dot \phi = u_\phi$, then the model can be represented as
\begin{equation}\label{eq:firstmodel}
    \begin{cases}
    \dot p_x &= v\cos(\theta)\\
    \dot p_y &= v\sin(\theta)\\
    \dot \theta &= v\frac{\tan{\phi}}{l}  \\
    %\textcolor{blue}{\dot v} &= \textcolor{blue}{u_v/m} \\
    \dot \phi &= u_\phi
    \end{cases}
\end{equation}
with the condition that $-\frac{\pi}{2} < \phi < \frac{\pi}{2}$ to stay in the approximation of no skid. We may consider  $-\phi_{max} < \phi < \phi_{max}$, with $\phi_{max}$, the maximum turn angle the wheels can rotate around their vertical axis. Usually, $\phi_{max} \approx \frac{\pi}{6}$.
\subsection{Rover kinematic equations}
If the control signal $u_\phi$ can modify the value of $\phi$ practically instantaneously (i.e., $\phi = u_\phi$), then we can consider for the Rover the following non-holonomic model,
\begin{align}\label{eq:model}
  \begin{cases}
  \dot p_x &= v\cos(\theta) \\
  \dot p_y &= v\sin(\theta)\\
   \dot \theta &= u_\theta
  \end{cases}
\end{align}
and the angle of the virtual front wheel can be computed as $\phi = \arctan\Big(\frac{lu_\theta}{v}\Big)$.

Eventually, we arrive at a simple unicycle model. Besides, we conducted experiments using small RC Rovers propelled by electrical motors, so we can take $v = u_v$, acting directly on the Rover's speed, disregarding inertial and resistance to advance.

\section{Parametric GVF for Rover guidance and path following}\label{sec:GVF_Rover}
Many algorithms for path following can be found in the literature. Among them, some of the most common are carrot chasing \cite{safwat2018}, line-of-sight \cite{gu2022}, or pure pursuit \cite{samuel2016}. A complete survey and comparison can be found in \cite{SurveyGVF}, where authors show using simulations that Vector Field (VF) is the technique that more accurately follows the path. We use the Guiding Vector fields methodology to guide the Rovers. At each point in space, the GVF provides a suitable heading direction for the Rover to converge towards the target path or remain on it when reached. This work is based on a  novel methodology, Singularity Free parametric GVF (SF-GVF), \cite{yao2021singularity},  which extends GVF to deal with self-intersecting paths and singularities (i.e. points at which the vector field is null) by defining parametric curves in an augmented space, where all singular points are removed.

Parametric GVF was designed to work with paths described by generic parametric equations in $\mathbb{R}^3$ and $\mathbb{R}^2$. In this last case: 
\begin{equation}\label{eqparf}
    \mathcal{P} := \{p_x = f_1(w),\ p_y =f_2(w)\},
\end{equation}
where $w$ is the parameter of the path, $p_x,p_y \in \mathbb{R}^2$ and $f_1,f_2 \in C^2$. Therefore, even though we are working with Bézier curves, any parametric curve could be used. 
\begin{figure}
    \centering
    \includegraphics[width=0.6\linewidth]{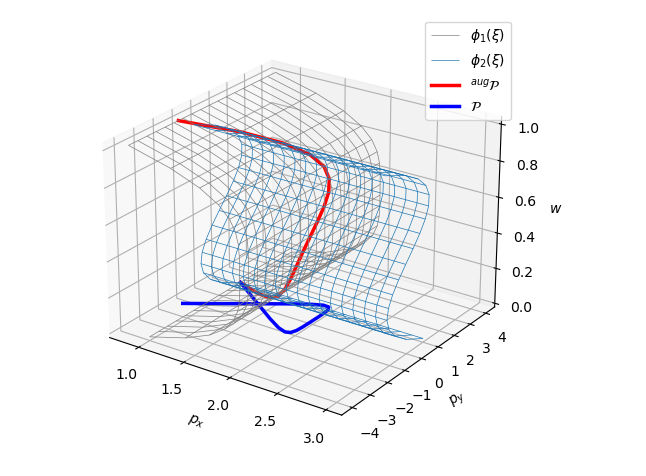}
    \caption{Illustration of the relationship among $\mathcal{P}$, $^{aug}\mathcal{P}$, $\phi_1(\xi)$ and $\phi_2(\xi)$. $\mathcal{P}$ is a Bézier's curve of degree 4, with control points $\mathcal{C} =\{(1,1),(3,-4),(1,0),(3,4),(1,-1)\}$}.
    \label{fig:bzaug}
\end{figure}

\subsection{Guidance problem}
The guiding vector field should be designed so that their integral curves converge to the desired path. Thus, if the vehicle aligns with the vector field, it will converge to the desired path. This approach is known as the vector-field guided path-following navigation problem; it can be briefly described as follows \cite{yao2021singularity}:
\begin{problem}[The vector-field guided path-following navigation problem]\label{pro1}
 Given a desired path $\mathcal{P} \subset \mathbb{R}^n$, to design a continuous differentiable vector field $\chi: \mathbb{R}^n \rightarrow \mathbb{R}^n$ such that the dynamical system $\dot{\xi}(t)=\chi(\xi(t))$ fulfills the following two conditions:
 \begin{enumerate}
     \item There is a neighborhood of $\mathcal{D} \subseteq \mathbb{R}^n$ such that for all $\xi(0) \in \mathcal{D}$ the distance between $\xi(t)$ and $\mathcal{P}$ approaches to zero as $t \rightarrow \infty$
     \item If a trajectory starts on the desired path, it remains in the desired path. $\xi(0) \in \mathcal{P}\Rightarrow \xi(t) \in \mathcal{P}, \forall t \ge 0$
 \end{enumerate}
 \qed
\end{problem}

To design an appropriate SF-GVF, we first define the following \emph{surfaces}, departing from the definition of the desired path $\mathcal{P}$ in (\ref{eqparf}), 
\begin{equation}\label{eqphis}
    \phi_1(\xi) = p_x - f_1(w),\ \phi_2(\xi) = p_y - f_2(w),
\end{equation}
where $\xi = (p_x,p_y,w) \in \mathbb{R}^3$ represents a vector on an  `augmented' space (i.e., the augmented space consists of the Rover position $(p_x,p_y)$ and the curve parameter $w$). We then define an `augmented path' as,
\begin{equation}\label{eqaugpath}
    ^{aug}\mathcal{P} :=\{\xi =(p_x,p_y,w) \in \mathbb{R}^3: \phi_i=0,\ i= 1,2 \}.
\end{equation}

Figure \ref{fig:bzaug} shows an example for a path $\mathcal{P}$ defined using a Bézier curve. Notice that $^{aug}\mathcal{P}$ is the intersection of the surfaces $\phi_1(\xi)$ and $\phi_2(\xi)$ and $\mathcal{P}$ is just the projection of $^{aug}\mathcal{P}$ onto $(p_x, p_y)$. The figure shows one of the main features of SF-GVF. The actual desired path $\mathcal{P}$ for the Rover (blue line) has a crossing point, and it is impossible to define a single vector for the Rover to follow to stay on the path at such a point. The augmented path $^{aug}\mathcal{P}$ instead unfolds $\mathcal{P}$ using the parameter $w$. Thus,  there is a single 3D vector for every point of the 3D unfolded curve. The Rover can now be guided using its 2D projection onto the plane $p_x, p_y$. 

We consider the function $e(\xi) = (\phi_1(\xi),\phi_2(\xi))^T: \mathbb{R}^3\rightarrow \mathbb{R}^2$ an error function, which gives a measurement of the splitting between a point $\xi$ and the path $\mathcal{P}$, $\lVert e(\xi) \rVert = 0 \Rightarrow \xi \in \mathcal{P}$, (where $||\cdot||$ is the Euclidean norm). 
We can now obtain a guiding vector field on the `augmented' space, departing from $\phi_1(\xi)$ and $\phi_2(\xi)$,
\begin{equation}\label{eqchiaug}
    ^{aug}\chi = \nabla\phi_1\times\nabla\phi_2 - \sum_{i=1}^2k_i\phi_i\nabla\phi_i.
\end{equation}

 The first addend of equation \eqref{eqchiaug} is perpendicular to $\nabla\phi_1$ and $\nabla \phi_2$. So, it is a vector tangential to the surfaces $\phi_1(\xi)=0$ and $\phi_2(\xi)=0$, pointing along $^{aug}\mathcal{P}$. It is called in \cite{yao2021singularity} the \emph{propagation term}. The sense of this propagation term can be changed just by swapping the gradients in the cross-product. The second addend points towards that surface (i.e., it is the normal component of the vector field) and, therefore, towards $^{aug}\mathcal{P}$. This term is called the \emph{converging term}, where $k_i>0$ are adjustable gains to fit the normal component of the field. 

The gradients $\nabla\phi_1=(1,0,-f'_1(w))^T$ and $\nabla\phi_2=(0,1,-f'_2(w))^T$ can be straightforwardly obtained, where $f_i'(w) := \dfrac{df_i(w)}{dw}$. Besides,
\begin{equation}\label{eq:propag}
\nabla\phi_1\times\nabla\phi_2 = (f'_1(w),f'_2(w),1).
\end{equation}

Thus, our `augmented vector field' is,
\begin{equation}\label{eqchiaug2}
    ^{aug}\chi(\xi) = 
    \begin{pmatrix}
    f'_1(w) - k_1\phi_1\\
    f'_2(w) -k_2\phi_2\\
    1+k_1\phi_1f'_1(w)+k_2\phi_2f'_2(w)
    \end{pmatrix}.
\end{equation}

Note that the third component of the vector in equation (\ref{eq:propag}) is always constant and equal to 1, regardless of the path's parametrization. Moreover, the terms $\nabla \phi_1 \times \nabla \phi_2$ and $\sum_{i=1}^2 k_i \phi_i \nabla \phi_i$ in equation (\ref{eqchiaug}) are orthogonal to each other and are linearly independent. As a result, the third component of $^{aug}\chi(\xi)$ is always non-zero. Therefore, $^{aug}\chi(\xi) \neq 0$ for all $\xi \in \mathbb{R}^3$. Thus, there are no singular points in the augmented field.

In summary, we have added a fictitious third coordinate to the dimensions of our problem, including the path parameter $w$ as a new variable. This has several advantages \cite{yao2021singularity}. First, it constructs a guiding vector field that leads the vehicle towards the path. Besides,  the augmented path is not self-intersecting, although the path could be, see figure \ref{fig:bzaug}, and the augmented GVF does not have singular points.

The vector field $^\text{aug}\chi$ defined in equation (\ref{eqchiaug}) is an effective solution to Problem \ref{pro1}, i.e. $\dot{\xi}(t)={}^{aug}\chi$ approaches the augmented path $^{aug}\mathcal{P}$ if $\xi(0) \notin {}^{aug}\mathcal{P}$ or remains in $^{aug}\mathcal{P}$ if $\xi(0) \in {}^\text{aug}\mathcal{P}$ . Formal proofs can be found in \cite[Theorem 3]{wkc2010} and \cite[Proposition 2, Theorem 2]{YAO2020108957}. Finally, in \cite[Theorem 2]{yao2021singularity} is proved that if $\mathcal{P}$ is parameterized as in equation (\ref{eqparf}), $\phi_1$ and $\phi_2$ are chosen like in (\ref{eqphis}) and there are no singular points in $^\text{aug}\chi$, the projected trajectory $\xi^p = (p_x,p_y)$ globally converges to the path $\mathcal{P}$.

 We will offer here a proof sketch of the exponential convergence of $^{aug}\chi(\xi)$ to $^{aug}\mathcal{P}$, according to the following proposition,

\begin{proposition}\label{prop:1}
    Let $\xi(t)$ be the solution to $\dot{\xi}(t) = \mkern 1mu ^{aug}\chi(\xi(t))$ with $^{aug}\chi(\xi(t))$ as defined in equation (\ref{eqchiaug2}), then $\xi(t)$ will converge to the augmented path $ ^{aug}\mathcal{P}$, defined in equation (\ref{eqaugpath}) as $t\rightarrow \infty$.
\end{proposition}

\begin{proof}
    The distance from the solution to the path at any time is defined as,
    \begin{equation}
        \text{dist}(\xi(t),\mathcal{P}) = \inf\{\Vert \xi(t)-p\Vert: p \in \mkern 1mu ^{aug}\mathcal{P}\}
    \end{equation}
    This distance can be approximated by $\Vert e(\xi(t)) \Vert$ because the norm of $e(\xi)=(\phi_1(\xi),\phi_2(\xi))^T$ is bounded and there are not singular points in $^{aug}\mathcal{P}$.
    Taking the $e(\xi)$ time derivative,
    \begin{equation}\label{eq:error}
        \dot{e}(\xi(t)) = N^T(\xi(t))\cdot \dot{\xi}(t) = N^T(\xi(t))\cdot \chi(\xi(t)) =- N^T(\xi(t))\cdot N(\xi(t)) \cdot K\cdot e(\xi(t)),
    \end{equation}
where $K = \text{diag}(k_1,k_2)$, $N=(\nabla \phi_1,\nabla \phi_2)$ and $N^T \cdot (\nabla \phi_1 \times \nabla \phi_2)=0$.

Now we define the following Lyapunov's function candidate:
\begin{equation}\label{eq:lyap}
    V(\xi(t)) = \frac{1}{2}e^T(\xi(t))\cdot K\cdot e(\xi(t)).
\end{equation}
One important property of $V(\xi(t))$ and the error vector norm $\Vert e(\xi(t))\Vert$ is,
\begin{equation}
    2\frac{V}{k_m} \geq \Vert e\Vert^2 \geq 2\frac{V}{k_M},
\end{equation}
where $k_m = \min\{k_1,k_2\}$ and $k_M = \max\{k_1,k_2\}$. We will omit $\xi(t)$ dependence for clarity. 
Taking the time derivative of the Lyapunov function,
\begin{equation}
    \dot{V} = \frac{1}{2}\left(\dot{e}^TKe+eK\dot{e}\right) = -\frac{1}{2}\left( e^TKN^TNKe+e^TKN^TNKe\right) = -e^TQe=-\Vert NKe\Vert^2
\end{equation}
 Notice that $Q = KN^TNK \succ 0$, $\det(Q) = k_1^2k_2^2\Vert \nabla \phi_1 \times \nabla \phi_2\Vert^2\geq0$.
(Recall equation (\ref{eqchiaug2})). Thus $\dot{V} = 0 \Rightarrow e(\xi) =0$.
Let $\lambda_{min}>0$ be the minimum eigenvalue of $Q$. Then,
\begin{equation}
    \dot{V}\leq -\lambda_{min}\Vert e\Vert^2 \leq -2\lambda_{min}\frac{V}{k_M}
\end{equation}
Thus,
\begin{equation}
\frac{k_m}{2}\Vert e \Vert^2 \leq V \leq V(e_0)\exp \left(-\frac{2\lambda_{min}t}{k_M} \right ) \leq \frac{k_M}{2}\Vert e_0\Vert^2   
\end{equation}
With $e_0 =e(\xi(0)$  the initial error. Eventually,
\begin{equation}
    \Vert e\Vert \leq \sqrt{\frac{k_M}{k_m}}\Vert e_0\Vert\exp \left(-\frac{\lambda_{min}t}{k_M} \right ) 
\end{equation}

\end{proof}

\subsection{Path following}
Once a dynamic model for the Rover, a desired path, and its GVF associated have been defined, we need to find a control law that makes the Rover trajectory converge to the desired path. First, we 
extend the dynamic model described in section \ref{sec:model}, equation \eqref{eq:model}, adding an equation for the new variable $w$:
\begin{equation}\label{eq:w_def}
\dot{w} = v\frac{\chi_3}{\sqrt{\chi_1^2+\chi_2^2}},    
\end{equation}
where $^{aug}\chi =(\chi_1,\chi_2,\chi_3)^T$ are the components of the `augmented' guiding field. The path parameter must adapt accordingly to the Rover's speed because the Rover will follow the guiding vector field evaluated at a specific $w$. As discussed in \cite[page 194]{weijiath}, the GVF path following process could also be considered as a special trajectory tracking algorithm \cite[page 506]{siciliano10}, in which the tracking parameter is state-dependent, closing a control loop. 

Therefore, the objective should be to align the Rover augmented velocity $\dot \xi = (\dot p_x, \dot p_y, \dot w)^T$ with the GVF, making it head in the direction of the field. Assuming the disturbances to the system can be neglected, we approximate the Rover velocity direction with its heading vector, $\hat{\dot p} \approx \hat{h} = (\cos(\theta), \sin(\theta))^T$, where $\hat\centerdot$ represents the normalization operator, and define $\chi^p$ as the projection of the guiding field $^{aug}\chi$ onto the plane $(p_x,p_y)$,
\begin{equation}
    \hat{\chi}^p := (\hat{\chi}_1,\hat{\chi}_2)^T = \frac{1}{\sqrt{\chi_1^2+\chi_2^2}}\begin{pmatrix}
        \chi_1\\ \chi_2
    \end{pmatrix}.
\end{equation}

\begin{figure}
    \centering
    \includegraphics[width=0.6\linewidth]{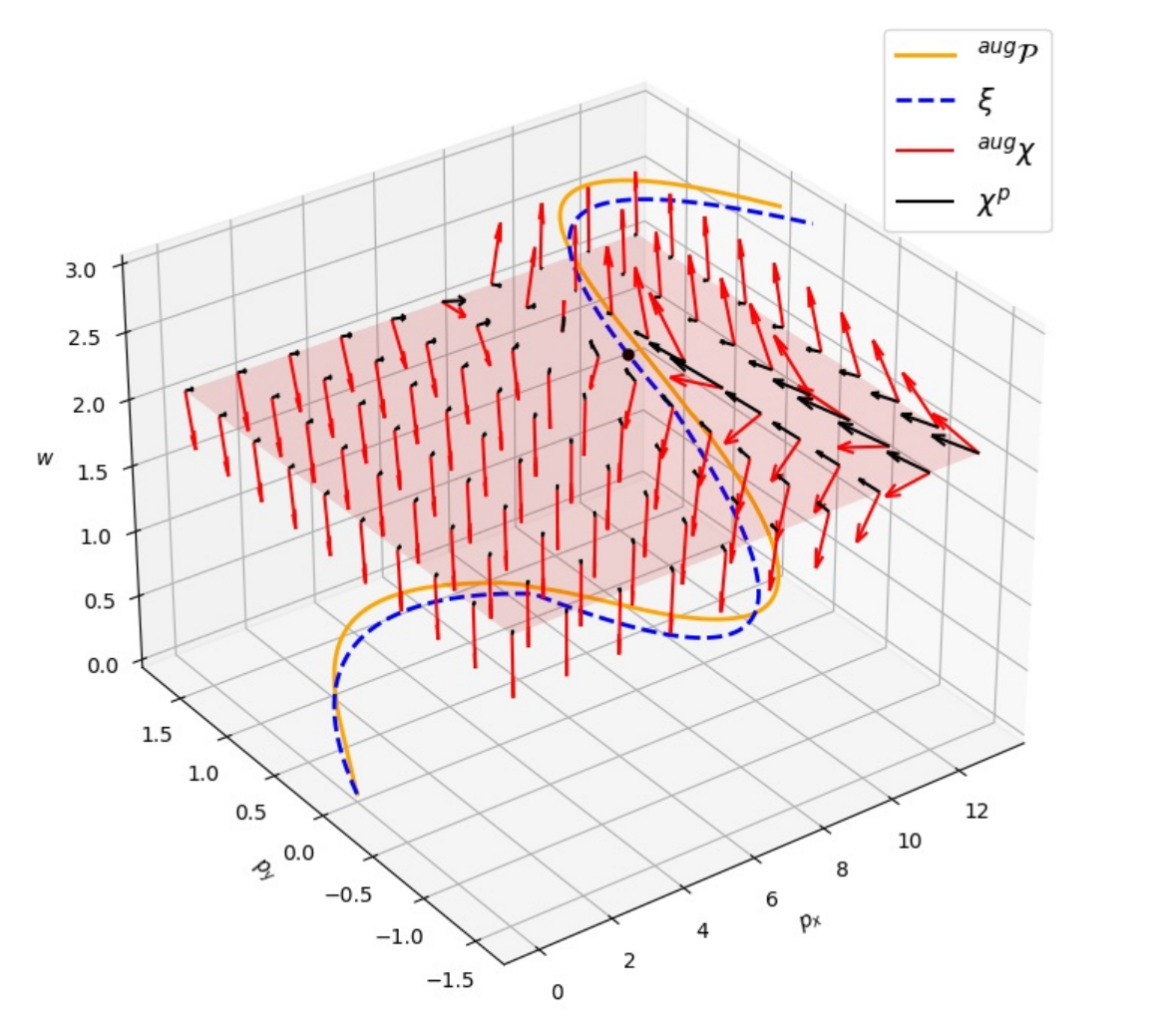}
    \caption{Illustration of the relationship between $^{aug}\chi$ and $\chi^p$. The vector field $\chi^p$ (shown as black arrows) is the projection of $^{aug}\chi$ (shown as red arrows) onto the plane $(p_x,p_y)$ of constant $w$ (shown as a black dot in the red plane). The solid orange line represents the augmented path, while the dashed blue line represents the augmented trajectory $\xi$. }
    \label{fig:chipa}
\end{figure}

Figure \ref{fig:chipa} shows the concepts relating to the augmented and projected vector field and trajectories. It is interesting to notice how the projected vector field changes following the parameter $w$.

Hence, let us consider the following error function between the Rover's orientation and $\hat{\chi}^p$:
\begin{equation}
e_* = \begin{pmatrix}
    \hat h - \hat{\chi}^p \\
    \dot w - v\frac{\chi_3}{||\chi^p||}
\end{pmatrix}
= 
\begin{pmatrix}
    \hat h - \hat{\chi}^p \\
    0
\end{pmatrix},
\end{equation}
which is $\mathbf 0$ if and only if $\hat h^T \hat{\chi}^p = 1$; therefore, the orientation is aligned with the vector field.
It is clear that thanks to the election made of $\dot{w}$ we only have to find a control law that makes $\hat{h} - \hat{\chi}^p \rightarrow 0$, i.e., a control law that asymptotically aligns the Rover with the projection of the augmented guiding field, $\chi^p$.
The variation of $\chi^p$ orientation as the Rover moves can be obtained as,
\begin{equation*}
    \frac{d}{dt}\hat{\chi}^p = \frac{1}{\Vert\chi^P\Vert}\left(I-\hat{\chi}^p\hat{\chi}^{pT}\right)\frac{d\chi^p}{dt} = \frac{1}{\Vert\chi^P\Vert}\left(I-\hat{\chi}^p\hat{\chi}^{pT}\right)J(\chi^p)v
\end{equation*}
Where $I\in \mathbb{R}^2$ is the identity matrix and $J(\chi^p)$ is the Jacobian matrix of the field $\chi^p = (\chi_1,\chi_2)$ with respect to $p = (p_x,p_y)$. Notice also that $\left(I-\hat{\chi^p}\hat{\chi}^pT \right) = E\hat{\chi}^p(E\hat{\chi}^p)^T$ is a projector in the direction orthogonal to $\hat{\chi^p}$. So,
\begin{equation*}
    \frac{d}{dt}\hat{\chi}^p  = -\left(\frac{\hat{\chi}^{pT}}{||\chi^p||}EJ(\chi^p)v\right)E\hat{\chi}^p = \dot \theta_d E\hat \chi^p.
\end{equation*}
Where $E$ is the 90 degree counter-clock rotation matrix $E = \left(\begin{smallmatrix}0 & -1 \\ 1 & \ \ 0\end{smallmatrix}\right)$ and we define:
\begin{equation}\label{eqthetapd}
\dot{\theta}_d := -\left(\frac{\hat{\chi}^{pT}}{||\chi^p||}EJ(\chi^p)v\right). 
\end{equation}
Notice that $\dot{\theta}_d$ is a scalar quantity, and it can be considered as the rotation rate of $\hat \chi^p$, since $E\hat{\chi}^p \perp \hat{\chi}^p$. Then, if the angular velocity of the Rover $\dot\theta$ converges to $\dot\theta_d$ while $\hat{h} \rightarrow \chi^p$, the Rover will follow the GVF and, eventually, will connect to the desired path. 
 A suitable Lyapunov function can be found in \cite{yao2021singularity} as $V = \frac{1}{2}||e_*||^2$, whose time derivative is:
\begin{equation*}
\begin{split}
    \dot V &= \left(\dot{\hat{h}}-\dot{\hat{\chi}}^p\right)^T \left(h-\hat{\chi}^p\right),\\
    \dot V &= \left(-\dot{\theta}\hat{h}^TE+\dot{\theta}_d\chi^{pT}E\right)\left(h-\hat{\chi}^p\right),\\
    \dot V &= (\dot \theta - \dot \theta_d)(\hat h^TE\hat{\chi}^p).
    \end{split}
\end{equation*}
Where $\dot{\hat{h}} = \dot{\theta} E \hat{h}$.
 Therefore, by choosing the control function
\begin{equation}\label{eq:ControlAction}
    \dot \theta = u_\theta = \dot \theta_d  - k_\theta\hat h^TE\hat{\chi}^p \; : k_\theta \in \mathbb{R}^+,
\end{equation}
 then $\dot V = -k_\theta(\hat h^TE\hat{\chi}^p)^2 \leq 0$. This means that $\hat h \rightarrow \hat{\chi}^p$ and $\dot \theta \rightarrow \dot \theta_d$ as desired. Notice that $k_{\theta}$ is an adjustable gain to modulate the control action.

We use spline segments with $ C^4$ continuity, while the endpoints connecting two curves have $ C^2$ continuity. This guarantees convergence, justifying using at least fifth-order Bézier curves.

\subsection{A note on field perturbation.}
We now add a perturbation to the vector field,
\begin{equation}
    \dot{\xi} = \chi(\xi(t)) + d(t)
\end{equation}
Where $d:\mathbb{R}_{\geq0}\rightarrow \mathbb{R}^2$ is a bounded function of time $t, \forall t\geq 0$. 
Then, the dynamics of the path-following error, equation (\ref{eq:error}), changes,
\begin{equation}\label{eq:ErrorWithdistirbance}
    \dot{e}(t) = N^T(\xi(t))\left(\chi(\xi(t)+d(t) \right)
\end{equation}
Using again equation (\ref{eq:lyap}) as Lyapunov's candidate function, its time derivative is,
\begin{equation}
    \dot{V} = \frac{1}{2}\left(\dot{e}^TKe+eK\dot{e}\right) = -\frac{1}{2}\left( (\chi+d)^TNKe+e^TKN^T(\chi+d) \right) = -\Vert NKe\Vert^2+d^TNKe
\end{equation}
and using Young's inequality, $d^TNke \leq \Vert NKe \Vert^2/2 + \Vert d\Vert^2/2$.
\begin{equation}
    \dot{V} \leq -\frac{1}{2}\Vert NKe\Vert^2+\frac{1}{2}\Vert d \Vert^2
\end{equation}
Following an identical reasoning as in proposition \ref{prop:1} proof, we arrive to,
\begin{equation}\label{eq:UpperboundError}
\dot{V} \leq -\frac{1}{2}\lambda_{min}\Vert e\Vert^2 + \frac{1}{2}\Vert d \Vert^2
\end{equation}
If $d(t)$ is bounded, $\Vert e \Vert$ will be bounded by $\sup_{t \in [0,\infty)}\Vert d(t) \Vert/\sqrt{\lambda_{min}}$, while if $d(t)$ vanishes as $t \rightarrow \infty$, then 
$||e(t)|| \rightarrow 0$ as $t \rightarrow \infty$.

\section{Bézier's curves for path planning}\label{sec:PathPlanning}
Whatever mission a final user wants to assign to an AMR involves the explicit or implicit definition of trajectories the vehicle should cover to fulfil its objectives. Probably the broadest extended method is the generation of Dubins' splines based on Dubins' curves \cite[page 880]{Dubin57,lavalle06}. For constant speed and predefined waypoints, a Dubins' spline is built by combining circle arcs and straight lines that minimize the time to cover the path. The method used a circle arc with the minimum radius allowed by the vehicle's speed. Another common approach is the use of N-splines. These are defined as $3^{\text{rd}}$ degree polynomials, which interpolate two consecutive waypoints of the desired path. The polynomial coefficients become entirely determined by imposing continuity for the splines and their first derivatives on the common waypoints. 

Even if SF-GVFs ensure a convergence to any $C^2$ continuous
parametric curve, not all parametric curves are suitable for an AMR mission.
While the previously mentioned curves are quite common in literature, they lack
several desired properties that ease the path creation and/or that allow to fit
the parametric curve to a real restricted area. That is, they usually lack: 1)
an intuitive adjustment of the waypoints by the operator, 2) analytical
knowledge of the area in which the curve is contained by a simple look at its
waypoints, 3) ease of computation,  and 4) ease of implementation in real
hardware. The first property allows for fast path creation, while the second for
an a-priori known area in which the curve will be contained. The last two
properties are related to hardware implementations, allowing the use of embedded
memory and computationally restricted systems. Bézier's curves meet all this
requirements, since 1) they are straightforward to adjust, 2) once we know the
waypoints of the Bézier curve, it is contained within its convex hull, so by
joining the points of the Bézier curve with straight lines, we have the area in
which the Bézier curve is contained; 3) they are easy to compute since they are
well-known and defined polynomials called Bernstein polynomials, and 4) for its
implementation only its waypoints are needed, and no previous computations must
be carried since its waypoints completely determine Bézier curves,
as they are the coefficients of the polynomials. Thus, they are very suitable
for AMR path following, as they can be easily defined and fit in tight areas,
and are computationally friendly for embedded systems.

Bézier curves have occasionally been used in robotics 
\cite{han2010bezier,hilario2011real,jolly2009bezier,kawabata2015path}, but no examples of their use have been found in the literature for the design of safe trajectories in uncrewed vehicles except for berthing \cite{yuan2023event}. Bézier curves allow the creation of a trajectory simply and intuitively \cite{simba2016real,hwang2003mobile}. In addition, they are quite suitable for working in combination with GVFs, as we will see later on. A brief description of them is given to expose their main characteristics.

We focus on two-dimensional Bézier curves, $f(w) \in \mathbb{R}^2$, of degree $n$. They are polynomials defined by a set of points, whose basis are the Bernstein polynomials $b_{k,n}(w) = \binom{n}{k}w^k(1-w)^{n-k}: w \in [0,1]$ where $\binom{n}{k}$ is the binomial coefficient. A degree  $n$ Bézier curve is defined as:
\begin{equation}\label{eq:Bézier}
   f(w) = (f_1(w),f_2(w)) =  \sum_{k=0}^n\beta_kb_{k,n}(w),
\end{equation}
where $\beta_k \in \mathbb{R}^2 : k \in [0,n]$ are the $n+1$ points that define the Bézier curve. Thus, as shown in equation (\ref{eq:Bézier}), there is no need to compute polynomial coefficients from the control points as, for instance, in standard basis cubic polynomials. This allows for less computations in the AMR vehicle, liberating resources for other tasks. Moreover, the points completely determine the shape of the curve, making it easier to adjust it manually, as in \cite{han2010bezier}.

The curve begins at point $\beta_0$ and ends at point $\beta_{n}$, called \textit{end points}, while  $\beta_k: 1 \leq k \leq n-1$, called \textit{control points}, are used to define the shape of the trajectory. For instance, two control points can shape the trajectory if a degree $n = 3$ Bézier curve is used. It's important to note that the Bézier curve is contained within the convex hull created by its points. This is useful when restricting the curve to a specific area. Hence, Bézier curves allow for an easy and intuitive way to fit the curve into a desired or particular shape or area by a) fixing the starting and end points and b) moving the control points to create a safe trajectory, knowing that it will be contained in its convex hull.
Finally, since Bézier curves are polynomials, an $n$ degree curve implies $C^{n-1}$ continuity, meaning that when $n \geq 3$, the acceleration, velocity and position are continuous. This property is necessary if curvature continuity is desired since it depends on the first and second derivatives of the parametric curve.
Bézier curves have many other properties that make them useful for various applications; for a complete description, we refer to  \cite{HANSFORD200275}.

Nevertheless, when generating paths, a single Bézier curve with many control points may be cumbersome for creating a desired shaped trajectory. Thus, a Bézier spline is designed for that purpose. When connecting two Bézier curves, the minimum requirement is to have $C^0$ continuity in the ending point $\beta_{n}^i$ of the curve $i$ and the starting point $\beta_{0}^{i+1}$ of the curve $i+1$ (except for the ending point of the last curve), that is, $\beta_{n}^i = \beta_{0}^{i+1}$ for every $ i \in \{0, 1, \hdots, N-2\}$, where $N$ is the number of Bézier curves that compose the spline. Moreover, if $C^1$ and $C^2$ continuity are desired, the following relationship between control points must hold
\begin{equation*}
    \begin{cases}
        {f^i}'(1) = {f^{i+1}}'(0) \Rightarrow \beta_{n}^{i} -  \beta_{n-1}^{i} =  \beta_{1}^{i+1}- \beta_{0}^{i+1}\Rightarrow\\ \beta_{1}^{i+1} \underset{C^1 \text{cont.}}{=} 2\beta_{n}^i - \beta_{n-1}^i \\
        {f^i}''(1) = {f^{i+1}}''(0) \Rightarrow \beta_{n}^{i} - 2\beta_{n-1}^{i} + \beta_{n-2}^{i} = \beta_{2}^{i+1} - 2\beta_{1}^{i+1} + \beta_{0}^{i+1} \Rightarrow \\ \beta_{2}^{i+1} \underset{C^1, C^2 \text{cont}.}{=} 4\beta_{n}^i - 4\beta_{n-1}^i + \beta_{n-2}^i
    \end{cases}
\end{equation*}
where ${f^i}' = \frac{df^i}{dw}$, $f^i(\cdot) = (f_1^i(\cdot), f_2^i(\cdot)) = \sum_{k=0}^n\beta_{k}^ib_{k,n}(\cdot)$ is the $i$th segment of the spline. This means that the first and second control points of the $i+1$ curve ($\beta_1^{i+1},\beta_2^{i+1}$) are also fixed and cannot be used to shape the trajectory. It can be shown that if $C^p$ continuity is desired for a Bézier spline of degree $n > p$, it implies that $p$ control points will be fixed, and only $n-1-p$ control points can be used to shape each segment of the spline. In our application, $C^2$ is necessary, and two control points would be enough. Therefore, $n=5$ degree Bézier curves are selected. 

A spline with $N$ segments and $C^2$ continuity can be expressed as
\begin{equation*}
    f_s(w) = 
    \begin{cases}
    f^0(w), & 0 \leq w \leq 1 \\
    f^1(w-1), & 1 \leq w \leq 2 \\
    \vdots & \vdots \\
    f^{N-1}(w-(N-1)), & N-1 \leq w \leq N
    \end{cases}.
\end{equation*}
where the continuity relations above hold for each Bézier curve. Note that $(w-i) \in [0,1]$, as required in the definition of Bézier curves.

Thus, since we will work with $N$ Bézier curves of degree $n=5$, each curve (i.e., each segment of the spline) has $C^4$ continuity and, at the points in which two curves connect there is $C^2$ continuity. The following reasons justify the usage of this type of spline. First, the guidance algorithm requires that the complete trajectory is a path with at least $C^2$ continuity. Second, if lower degree curves were used (i.e., $n=3,n=4$) and $C^2$ continuity is required, this would imply that less than two control points could be used to shape each segment of the spline, losing the ease of creating Bézier curves. Finally, only one segment and a higher degree Bézier curve could be used for the complete trajectory. However, in contrast to fixing ending points and only changing the $i$ and $i+1$ segment when moving a control point, creating a single polynomial leads not only to a complex adjustment of the curve to the desired area but to a higher-order polynomial computation, which could lead to lack of numerical stability \cite{numericalstabilitybezierBspline}.

\section{Speed Control Problem}\label{sec:speed_control}
Several works can be found in the literature on adapting guidance controllers to the curvature, like in \cite{Leng17}, for guidance algorithms to achieve better convergence. A speed controller for an AMR vehicle is helpful when following curvature-changing paths. This allows us to reach a constant speed setpoint and adapt the speed to path characteristics. The non-varying setpoint provides for constant speed path-following while changing the setpoint depending on some information about the desired path (e.g., curvature) should speed up the convergence of the guidance controller. This approach is simple because only the desired reference speed changes; thus, whatever speed controller can be used, for example, in \cite{cao2019bio}, a finite time double sliding surface guidance algorithm is used for Subway's speed curve tracking.

In recent years, there has been increased attention towards determining trajectory-dependent speed profiles for guidance due to the rise of autonomous road vehicles \cite{gamez2017dynamic}. Most works use a-priori information about the desired trajectory to compute the path curvature and use this information to adapt vehicle speed, preventing slipping and rollover. Trajectory information can be extracted from maps, like \cite{villagra2012smooth} that used a closed-form speed profiler combined with the path planner, providing a continuous velocity reference. Or \cite{Li2012AutomatedIA} that obtains the curvature using the Geographic Information Systems (GIS) roadmaps. In other cases, certain assumptions are made about the curvature of roads to adapt the vehicle's speed, as in \cite{park2015}. 
  
\subsection{Speed controller}
To control the Rover speed, we use as a control signal $u_v$ (i.e., the throttle) the output of a feedforward + PID controller:
\begin{equation}\label{eq:feedPI}
 u_v = k_fv_{ref} + k_pe_v(t) + k_d\dot e_v(t) + k_i\int_{0}^{t}e_v(\tau)d\tau,
\end{equation}
where $v_{ref} \in \mathbb{R}_{\geq 0}$ is the desired speed, $e_v(t) = v_{ref}-v(t)$ is the error between this one and the actual speed of the Rover, and $k_f,k_p,k_d,k_i \in \mathbb{R}_{\geq0}$ are, respectively, the feedforward, proportional, derivative and integral control constants. The feedforward controller allows reaching a constant speed depending on $v_{ref}$, while the PID controller acts in the presence of error, allowing the desired speed to be reached without steady-state error.

Recall that, as was pointed out in section \ref{sec:model}, we have ignored the dynamic of the Rover. Thus $v \sim \propto u_v$ and we can consider,
\begin{equation}
v = k_fv_{ref} +k_p(v_{ref}-v)+k_d(\dot{v}_{ref}-\dot{v})+k_i\int_0^t(v_{ref}-v)dt
\end{equation}
Taking time derivatives, considering a constant $v_{ref}$ and  regrouping terms, we get a linear system,
\begin{equation}
\begin{pmatrix}
\ddot{v}\\
\dot{v}
\end{pmatrix} = 
\begin{pmatrix}
-\frac{1+k_p}{k_d}&-\frac{k_i}{k_d}\\
1&0
\end{pmatrix}
\begin{pmatrix}
\dot{v}\\
v
\end{pmatrix}+
\begin{pmatrix}
1\\
0
\end{pmatrix} \frac{k_i}{k_d}v_{ref}.
\end{equation}
It is immediate to see that the system matrix always has negative spectral abscissa, i.e. the system is stable and $v\rightarrow v_{ref}$ as $t\rightarrow \infty$, due to the control integral term.

Note that if the setpoint change is slower than the controller's time constant, then the vehicle will follow the desired time-changing speed setpoint. This plays a crucial role in the next section since we will use a smooth (i.e., differentiable) and variable setpoint that depends on curvature.

In our case, the Rover measures its position directly from the GPS, discarding data from inertial sensors. Our Rovers are small outdoor vehicles subject to bumpy and stony terrain and several other factors. This bounciness affects the accelerometers of the IMU, adding a significant amount of noise, almost equal to or larger than the actual acceleration of the Rover due to displacement. Besides, the Rover's acceleration transient time is faster than the noise perturbations (due to its relatively small mass). Therefore, the acceleration due to an actual increase or decrease in speed could not be differentiated from the noise signal. In addition, a proper measure of the noise process variance is complex since it depends on the terrain.

Thus, the GPS speed is critical, but this signal is also noisy. Filtering is needed to allow for the operation of a speed controller (especially the derivative controller). For this purpose, we use a simple \emph{moving average} filter. Therefore, our Rovers measure the speed using only the GPS signal filtered with a \emph{moving average} filter, and they use this filtered speed measurement $v$ as an input to the speed controller in equation \eqref{eq:feedPI}.
\subsection{Changing speed setpoint}
Typically, path-following algorithms consider a constant speed when following a specific curve. This speed is chosen so that the vehicle can follow the trajectory with sufficiently small (guidance) control inputs. Moreover, their heading cannot change instantly when dealing with non-holonomic vehicles like the model in equation (1). This means that the radius of curvature that it can trace increases (respectively decreases) as speed increases (decreases). Therefore, if the speed is held constant, it must be restricted to the maximum value that ensures the Rover can follow the trajectory. That is, the speed of the Rover must allow its guidance controller to converge to the trajectory, regardless of high curvature segments. That is because guidance control signals (i.e., inputs to the steering angle) at relatively high speeds can cause a rollover. Moreover, the saturation of the steering angle creates the same problems explained above. 

Besides, there are applications in which a constant speed is not desirable or interesting. For example, if the trajectory must be completed below a certain time stamp, an interesting approach is that the vehicle accelerates where the curvature is low or zero (i.e., straight lines) and reduces its speed when curvature increases. For this purpose, we propose that the speed setpoint can be changed according  to the path's curvature, following the next equation,
\begin{equation}\label{eq:setpoint}
    v_{ref}(\kappa) = (v_{max}-v_{min})\exp({-c_{\kappa}\kappa^2}) + v_{min},
\end{equation}
where $v_{max}$ and $v_{min} \in \mathbb{R}_{\geq 0}$ are the maximal and minimal Rover speed ($v_{max} \geq v_{min}$), $\kappa \in \mathbb{R}$ is the curvature and $c_{\kappa} \in \mathbb{R}_{\geq 0}$ is a constant used to penalize the speed as curvature increases. Since we are dealing with parametric curves, the curvature formula is well known \cite{ModernGeometry} and is defined as
\begin{equation}\label{eq:curvature}
  \kappa(w) = \frac{ f_1'(w) f_2''(w) -  f_1''(w)f_2'(w)}{( f_1'(w)^2 +  f_2'(w)^2)^{3/2}} .
\end{equation}
Thus, the curvature is computed using the parameter of the path $w$, that is, equation \eqref{eq:curvature} calculates the curvature of the point at which the vector field is pointing to (see figure \ref{fig:pprz_gcsField}).

Assuming that $c_\kappa$ is greater than zero in equation \eqref{eq:setpoint}, we can observe that the speed setpoint, which will be utilized in the controller \eqref{eq:feedPI}, is limited by $v_{ref} \in [v_{min}, v_{max}]$. The maximum value of the setpoint is achieved when $\kappa \rightarrow 0$, while the minimum value is obtained when  $\kappa \rightarrow \infty$. By adjusting $c_\kappa$ appropriately, the speed of the Rover can be slowed down when it approaches areas with relatively high curvature. It's interesting to note that the exponential argument is quadratic. This solves two problems: 1) $\kappa$ can take negative values depending on the sign of the curvature, and 2) taking $\kappa^2$ instead of $|\kappa|$ makes the setpoint differentiable, providing the curve has at least $C^3$ continuity to have a differentiable curvature, which is a fine property if, for example, a derivative or nonlinear controller is desired.

In summary, the Rover will reach its maximum speed whenever it is following a straight path, $\kappa = 0$; it will get a stable speed $v_{max}>v \geq v_{min}$ if it is following a curve of constant $\kappa$, i.e. a circle or circle section and it will accelerate o decelerate if $\kappa$ is decreasing or increasing. Parameters in equation \ref{eq:setpoint} allow us to tune the WMR behaviour to the path and terrain features.

\section{Results}\label{sec:results}

\subsection{Experimental Platform}
 We use the same software for simulations and experiments. In this way, the algorithms implemented for the simulation can also be used in the experiments without any further change. This section briefly describes the hardware and software platforms used for simulations and experiments.
 
\subsubsection{Rover's Hardware}
\begin{figure}[t]
    \centering
    \subfloat[\centering View of the team of Rovers.]{{\includegraphics[width=0.45\linewidth]{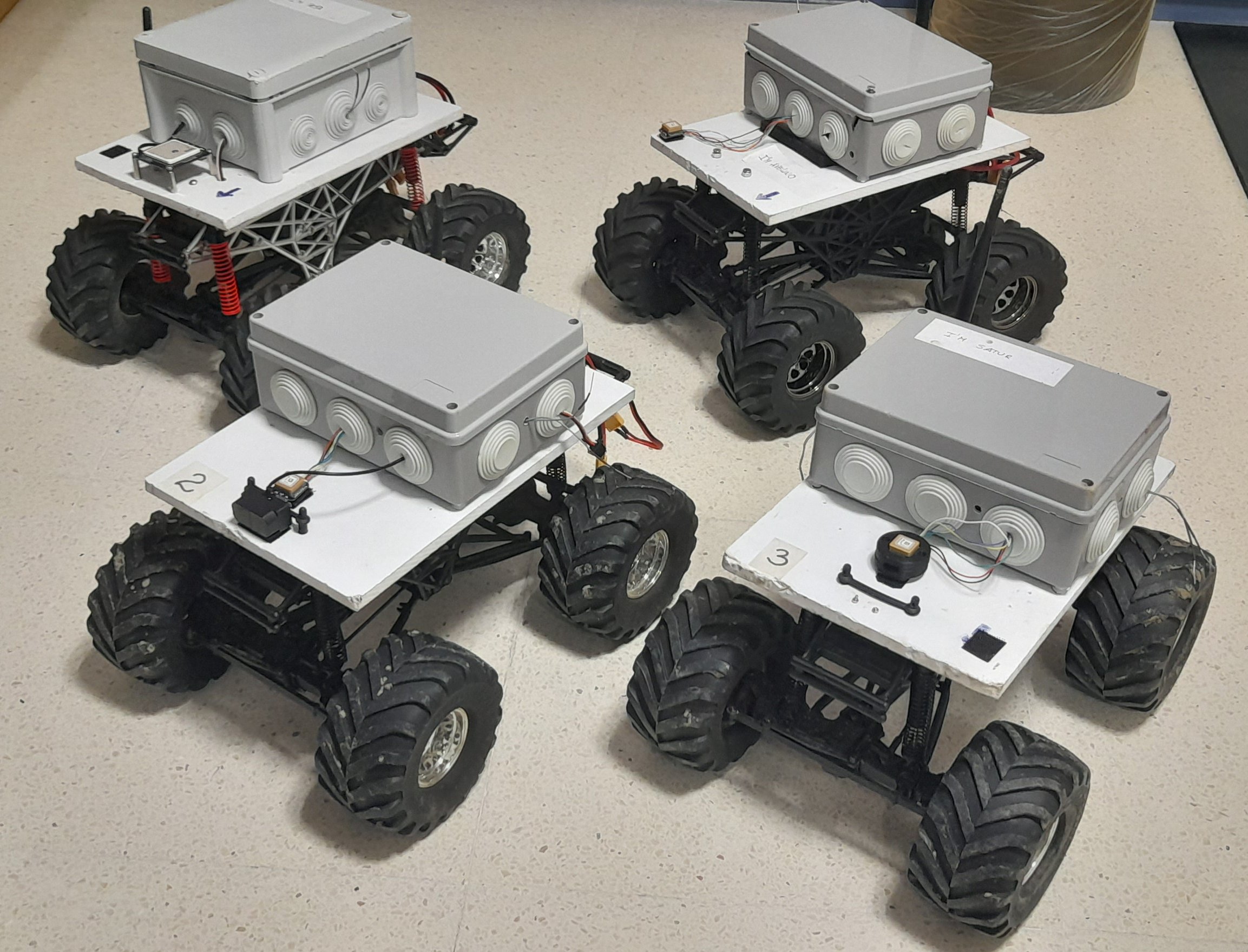} }}%
    \qquad
    \subfloat[\centering Overview of the mechanical system of the Rovers.]{{\includegraphics[width=0.45\linewidth]{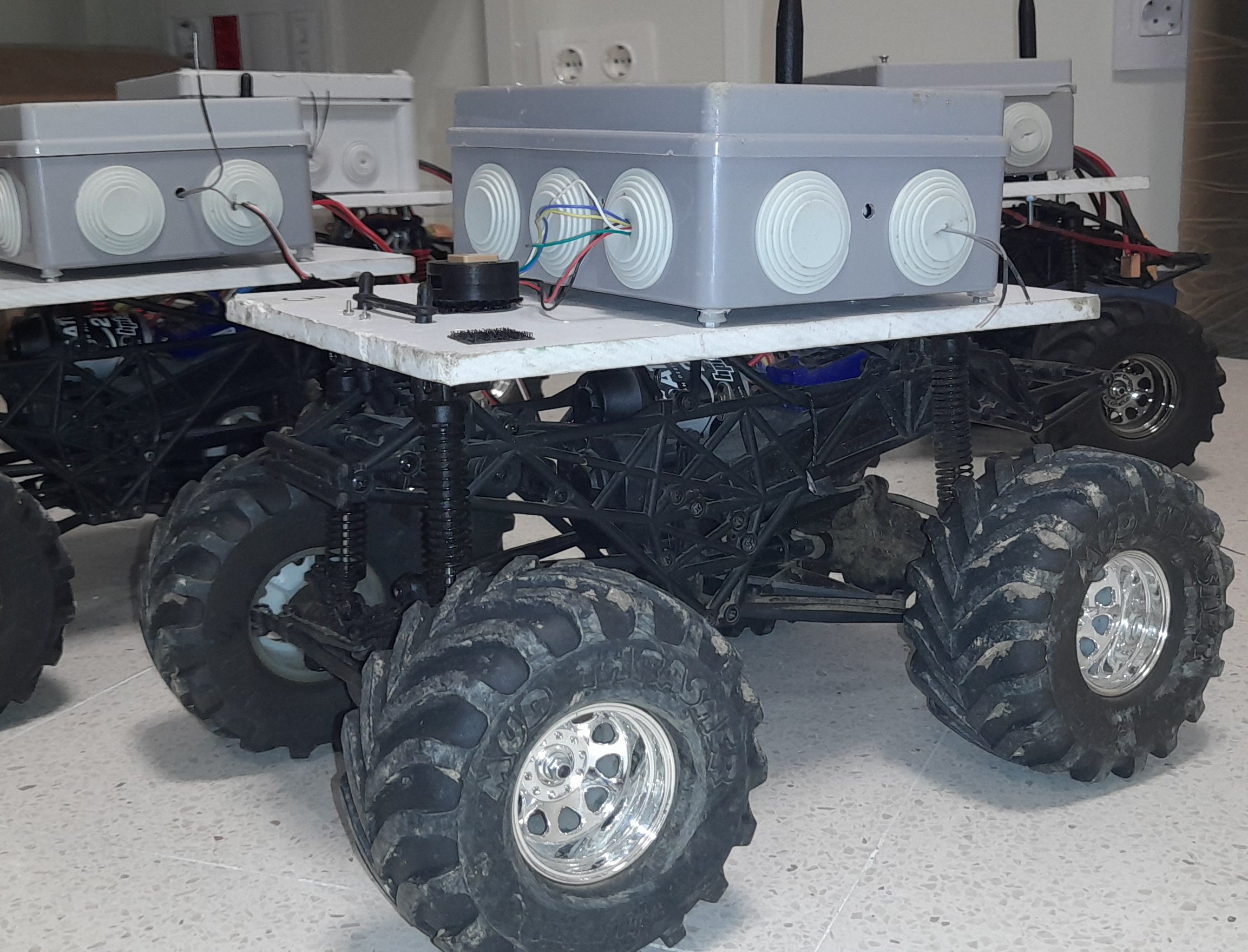} }}%
    \caption{Rovers}%
    \label{fig:RoverTeam}%
\end{figure}
\begin{figure}
    \centering
    \includegraphics[width=0.6\linewidth]{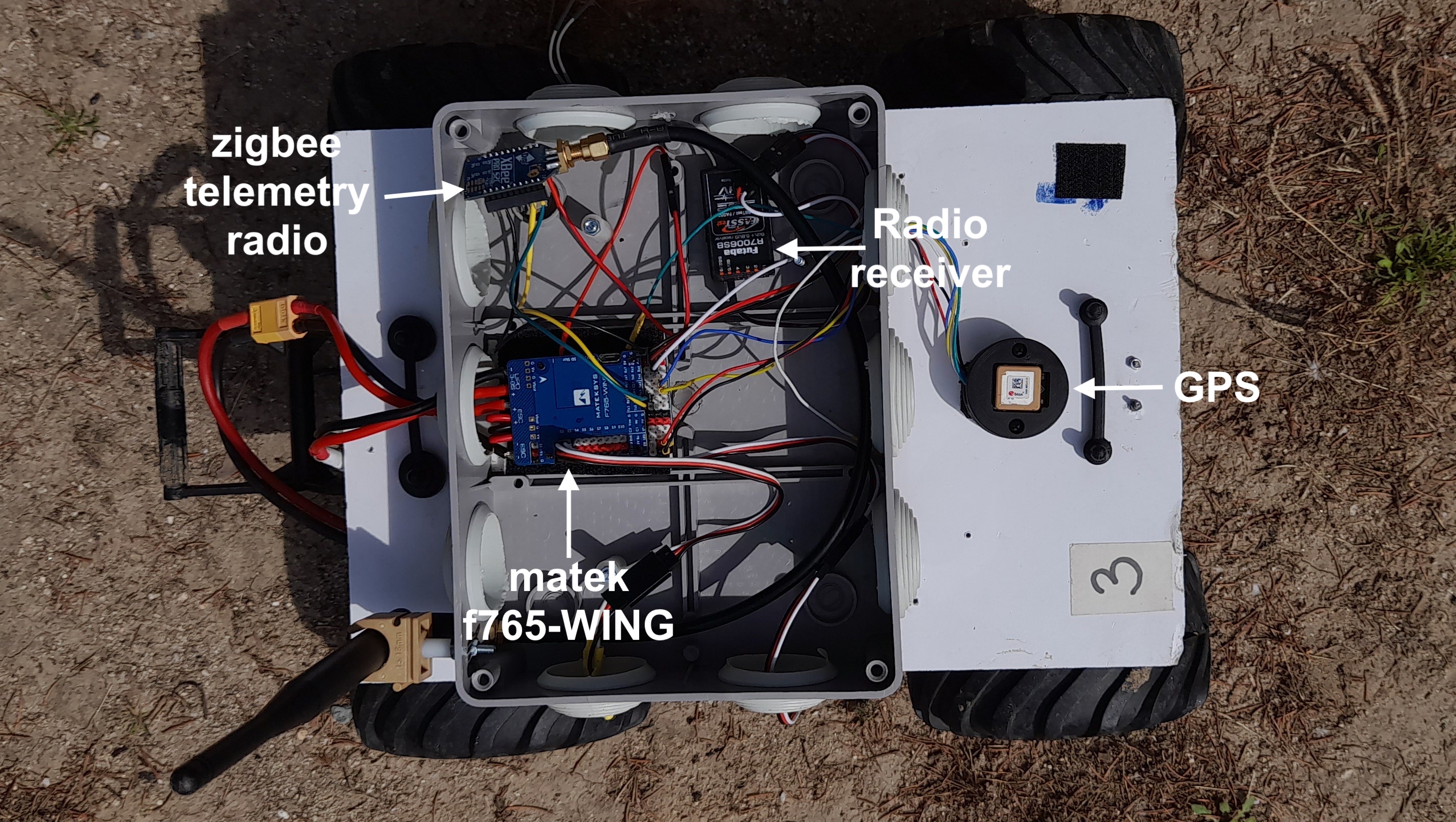}
    \caption{Rover's electronics hardware setup in an outdoors experiment.}
    \label{fig:Rover_hardware}
\end{figure}
Figure \ref{fig:RoverTeam} shows several of our Rovers.
These Rovers are commercial RC electric cars modified to provide a robust mechanical structure for their use as a testing platform, not only for algorithms designed for these vehicles but also for more complex systems, such as Unmanned Surface Vehicles (e.g., a boat whose direction is controlled by a rudder or a differential drive vessel) or fixed wing planes. The dimensions of the Rover are collected in table \ref{tab:Roversmeasurements}.
We have removed the commercial radio receiver from the Rovers and mounted on top of it a {PVC foam board} and a potting box as shown in figure \ref{fig:Rover_hardware}. These Rovers have a front-wheel Ackerman-type steering, which mechanically coordinates the angle of the two front wheels mounted on a standard axle for steering. Moreover, they are all-wheel drive and equipped with differential gears in both axes, reducing the skidding when turning since it allows each wheel (in each axis) to rotate at a different speed.
\begin{table}[h]
\centering
\begin{tabular}{|c|c|c|c|c|c|c|}
\hline
\begin{tabular}[c]{@{}c@{}}Wheelbase \\\end{tabular} &
  \begin{tabular}[c]{@{}c@{}}Axle Track \end{tabular} &
  Length &
  Width &
  Height &
  \begin{tabular}[c]{@{}c@{}}M.S.A \\ outer wheel \end{tabular} &
  \begin{tabular}[c]{@{}c@{}}M.S.A \\ inner wheel\end{tabular} \\ \hline
25 cm &
  23 cm &
  40 cm &
  30 cm &
  29 cm &
  15 degrees &
  10 degrees \\ \hline
\end{tabular}
\caption{Rover's physical measurements. MSA stands for Maximal Steering Angle.}
\label{tab:Roversmeasurements}
\end{table}
The Rovers' movement is controlled by two motors: (1) an HPI Racing Saturn brushed DC motor that provides thrust to all wheels and (2) a servo motor that drives the steering. 
The onboard autopilot is a Matek f765-Wing flight controller whose Microcontroller Unit (MCU) is the STM32F765VIT6 from ST Microelectronics. Although it has been designed to control fixed-wing drones, the autopilot suits our purposes well, supplying a common platform for different types of autonomous vehicles. In addition to a built-in Inertial Measurement Unit (IMU) in the flight controller, the Rovers have several devices connected to it: 1) an ublox GPS (SAM-M8Q) receiver with a built-in compass, 2) a Digi Xbee S2C RF transceiver that allows both monitoring the Rovers' parameters and sending control signals to them from a ground station (see below) and 3) a Futaba RC receiver with SBus protocol, that allows the Rovers to be remotely controlled using a RC Futaba radio and transfer the control to and from the autopilot. These devices also grant access to the necessary states: the position  $(p_x,p_y)$ and attitude $\theta$, relative to the inertial frame $W(p)$, needed for the guidance controller shown in equation \eqref{eq:ControlAction}, as well as access to the speed $v$, used (after filtering) by the speed controller in equation \eqref{eq:feedPI}. A 7.4 V Li-Po Battery powers the complete system with a capacity of 3000 mAh.

\begin{table}

    \centering
    %\resizebox{\textwidth}{!}{% 
    \begin{tabular}{|c|c|c|} \hline
        Brushed DC motor            & HPI Racing Saturn                         & Provides thrust to all wheels\\ \hline
        Servo motor                 &                                           & Drives the steering\\ \hline
        Microcontroller unit        & STM32F765VIT6 (ST Microcontrollers)    & Onboard Matek f765-Wing autopilot\\ \hline
        Inertial Measurement Unit   & MPU-6000 and ICM20602 (InvenSense)     & Onboard Matek f765-Wing autopilot\\ \hline
        GPS                         & ublox GPS (SAM-M8Q)                       & Provides speed and positioning  \\ \hline
        RF transceiver              & Digi Xbee S2C RF                          & Provides communication to ground station\\ \hline
        Radio receiver controller   & RC Futaba with Sbus protocol              & Allows to remote control the rover \\ \hline
        Battery                     & 7.4 V Li-Po 3000 mAh                      & Rover power\\ \hline
    \end{tabular}
    %}
    \caption{Rover's hardware components}
    \label{tab:rov_hardware}

\end{table}

\subsubsection{Paparazzi UAV environment}
As noted in section  \ref{sec:intro}, we use Paparazzi as a development environment to program autonomous vehicles. Paparazzi is a free and open-source hardware and software project intended to create a flexible autopilot system. Although it was initially designed to deal with unmanned aerial vehicles (UAVs), researchers from different universities have extended its applications to other autonomous vehicles. Moreover, Paparazzi supports
multiple hardware designs. To do so, it includes 1) a cross compiler, allowing the user to compile code for any supported flight controller board, 2) an uploader tool to upload the implemented code to the flight controller, and 3) support for several sensors and devices. The reader interested may address to Paparazzi wiki pages: \url{https://wiki.paparazziuav.org/wiki/Main_Page}. 

\begin{figure}
    \begin{minipage}[c]{0.45\linewidth}
    \includegraphics[width=\linewidth]{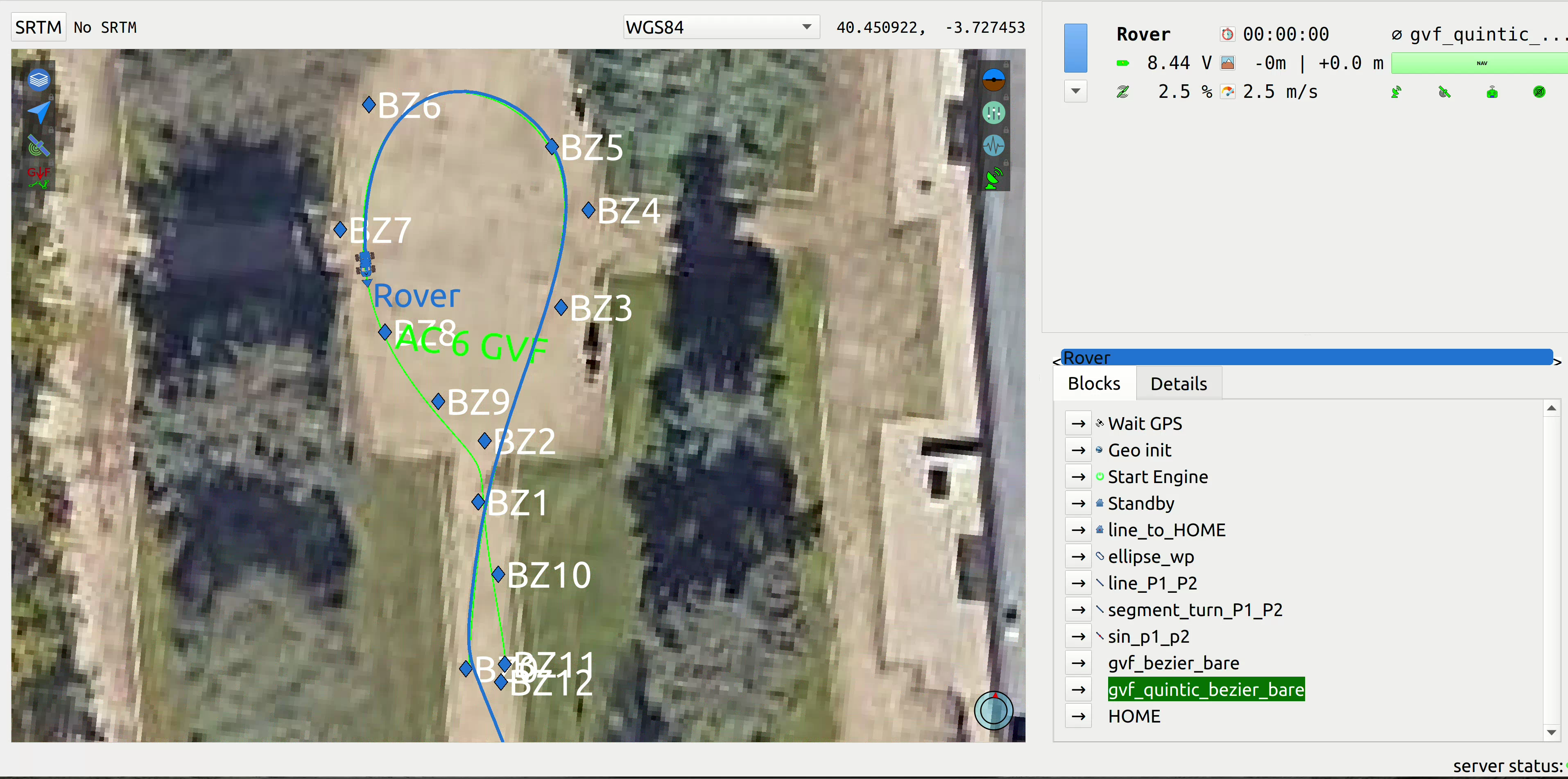}
    \caption{Screenshot from the Paparazzi's GCS in simulation. Bézier points are represented as blue diamonds, while the desired trajectory is represented as a green line. The actual trajectory traced by the Rover is shown in blue.}
    \label{fig:pprz_gcsNoField}
    \end{minipage}
    \hfill
    \begin{minipage}[c]{0.45\linewidth}
    \includegraphics[width=\linewidth]{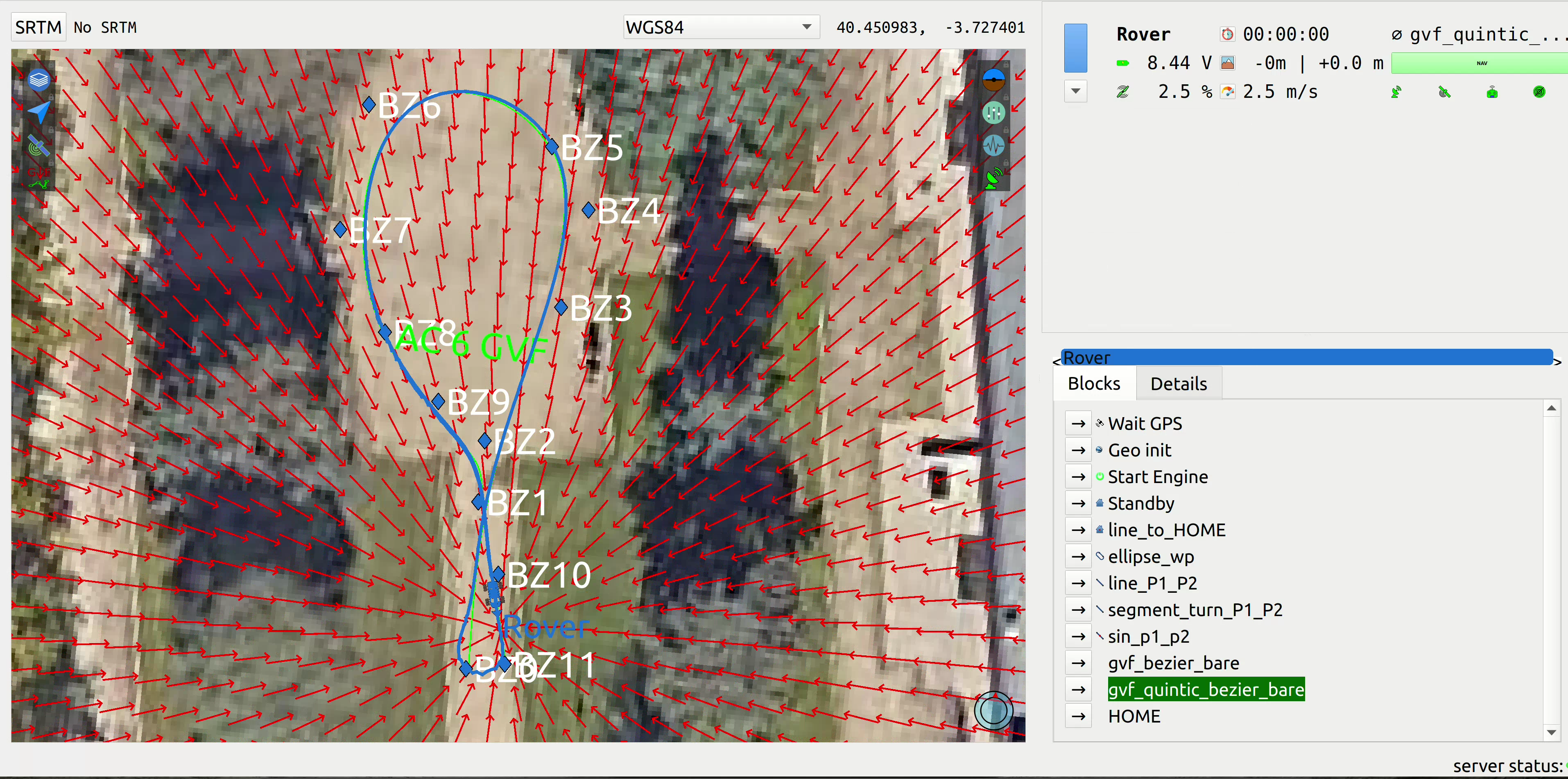}
        \caption{Screenshot from the same simulation as figure \ref{fig:pprz_gcsNoField} but forwarded in time and showing the vector field $\hat{\chi}^p$ as red arrows.}
    \label{fig:pprz_gcsField}
    \end{minipage}%
\end{figure}

Paparazzi also provides a simulator with different simulation vehicle models (e.g., the Rover's model, fixed-wing aircraft models and rotorcraft models) as well as sensor models. Besides, it is possible to modify the models or add new ones to fit the model to a specific AMR. The simulator is intended to test the performance of new algorithms before implementing them in the actual vehicles, saving time and resources.

The third and last component of Paparazzi is a Ground Control Station (GCS),  shown in figures \ref{fig:pprz_gcsNoField} and \ref{fig:pprz_gcsField}, where agents' positions are visualised, and different sensor and internal variables can be monitored.  The Paparazzi GCS is coded in C++ and can be customised to include new properties. For instance, we have added the code to define, show, and modify the Bézier curves used for path planning on the fly.
To offer maximum flexibility and openness, the
Paparazzi ecosystem was designed from the start as a modular one  \cite{gati2013open}. The user can specify the type of vehicle and sensors used, the flight plan, and the desired telemetry data using Extensible Markup Language (XML) files. This allows the user to choose between several XML modules without developing any C-Code. Therefore, thanks to the modular software architecture, generating the basic navigation code for the Rovers has been accelerated by using the different modules available in the paparazzi framework, such as GPS, radio control, wireless communication and others.

\begin{figure}[ht]
    \centering
    \includegraphics[width=\linewidth]{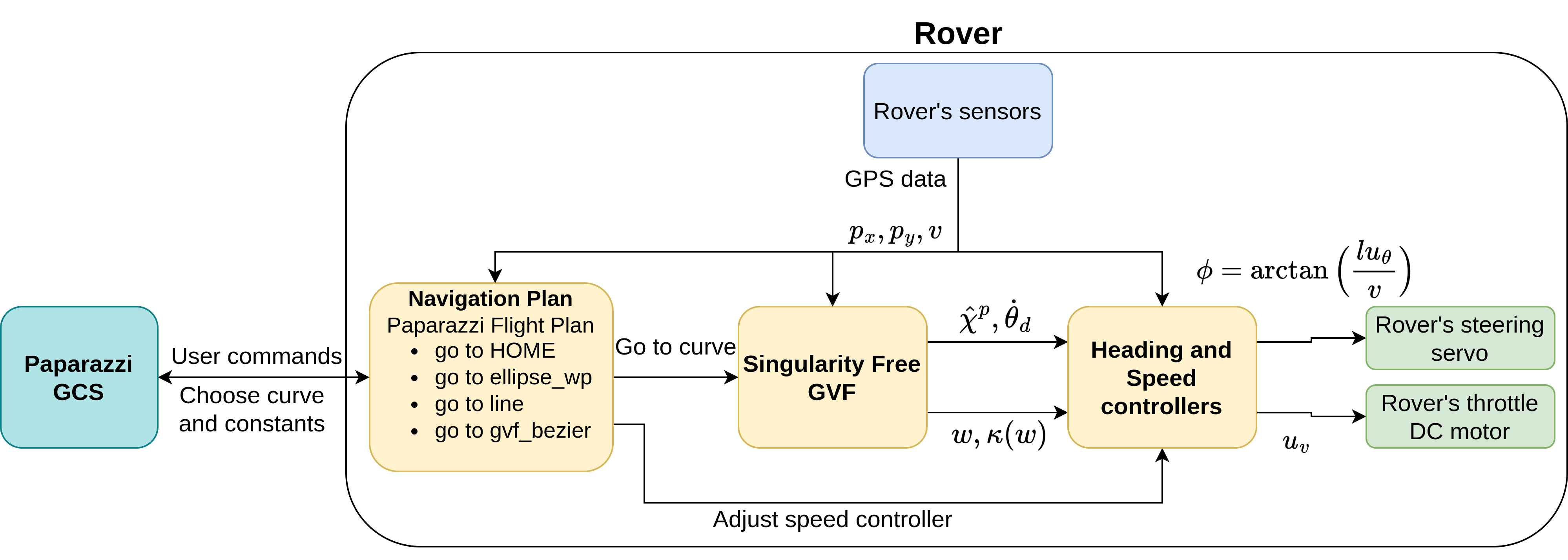}
    \caption{Schematic representation of the system. Once the code has been uploaded to the AMR from Paparazzi, GCS commands can be given to the Rover via telemetry. Then, the Rover follows the trajectory and speed setpoint according to the modules shown in the figure.}
    \label{fig:RoverSchema}
\end{figure}

Figure \ref{fig:RoverSchema} shows a block diagram with the system's main features. Using the Paparazzi GCS, the user can give real-time commands as inputs to the Rover's autopilot. Examples of inputs include path following commands of a desired curve --waypoints, speed setpoints, and guidance and speed controller constants. The rest of the system runs in the Rover's flight controller, allowing for a fully autonomous mode. Concerning the autonomous mode, states $(p_x,p_y)$ and speed $v$ are obtained from GPS data and sent to the SF-GVF module. This module computes the desired trajectory $f(w)$, the vector field $\hat \chi ^p$ and the term $\dot \theta_d$ needed for $u_\theta$, as well as $w$ and $\kappa(w)$ required in the speed controller. The heading and speed controllers generate the control signals $u_\theta \rightarrow \phi$ and $u_v$ by using the states, guidance information from the SF-GVF, and controller constants given by the user via Paparazzi GCS. Moreover, the states are fed to the Navigation Plan block to show these variables in real-time in paparazzi GCS. Finally, the control signals are sent to the lowest level of the Paparazzi UAV (not shown in the figure; it maps the values of $\phi$ and $u_v$ to PWM signals) to turn the Rover's steering servo and spin the DC motor.

The heading and speed controllers have been implemented in Paparazzi (C-code), allowing us to use them in simulation and in the experimental platform. First, the algorithm in section \ref{sec:GVF_Rover} has been modified to be used with Rover vehicles since it was already implemented in Paparazzi for fixed-wing aircraft. Moreover, the capability of following Bézier curves has been added. Thus, the user can not only create a desired Bézier curve in the Paparazzi GCS but modify it when desired (e.g., during an experiment) since it is the flight controller in the Rover that automatically computes the Bézier curve control points for continuity conditions.
Second, the speed controller of equation \eqref{eq:feedPI} has been implemented using the modules Paparazzi provides. Third, the changing speed setpoint of section \ref{sec:speed_control} and a moving average filter for the measured speed have been implemented. 

Finally, to see in real-time the desired parametric curve $f(w)$, the authors have implemented the code to show the trajectory and also the varying vector field $\hat \chi^p$, as shown in figures \ref{fig:pprz_gcsNoField} and \ref{fig:pprz_gcsField} (where a fifth order Bézier curve with $C^2$ continuity is shown). This allows us to adjust the guidance constants (i.e., $k_1,k_2$ and $k_\theta$) by visualising the deviations from the desired trajectory and the point the vector field is pointing to. 

It is important to note that waypoints can be adjusted to fit the trajectory into the desired area. By simply moving the desired control or endpoints, the user can shape the path as they wish. Onboard software recalculates the trajectory whenever a point has changed. This allows for considerable flexibility in experimental environments where uncertainties in position may arise.

The main code for following third-order Bézier curves is under the central paparazzi repository: \url{https://github.com/paparazzi/paparazzi}, for following both third-order and fifth-order Bézier curves is under the repository \url{https://github.com/alfredoFBW/paparazzi_alfredoFBW}, while the code to visualise both Bézier curves and the vector field in paparazzi GCS is under  \url{https://github.com/alfredoFBW/PprzGCS}.

\subsection{Simulation results}
A trajectory is created to test the SF-GVF and heading controller in simulation using $N = 3$ fifth order ($n = 5$) Bézier curves. As explained in section \ref{sec:PathPlanning}, a degree fifth Bézier curve is defined by $n+1=6$ points: the starting and end points and four control points, which can be used to shape the curve. Since the spline has $C^2$ continuity, the endpoint of the segment $i$ is the starting point of segment $i+1, i  \in [0,N-2] $, and the two control points  $\beta^i_{1}, \beta^i_{2}$ of segment $i$ came defined by the values of the second and third to last points of the previous segment $\beta^{i-1}_{4}, \beta^{i-1}_3$, giving us two control points to shape each spline segment. Thus, we have $3(N + 1) = 12$ configurable points for the complete spline, which we define as $\beta^s_k$. In particular, $\beta^s_k: k \in\{0,5,8,11\}$ define each segment's starting and end points, while the rest are used as control points to modify each segment's shape independently. Therefore, the curve starts at $\beta^s_0$ and ends at $\beta^s_{3N-2} = \beta_{11}$.
\begin{figure}[ht]
    \centering
    \includegraphics[width=0.7\linewidth]{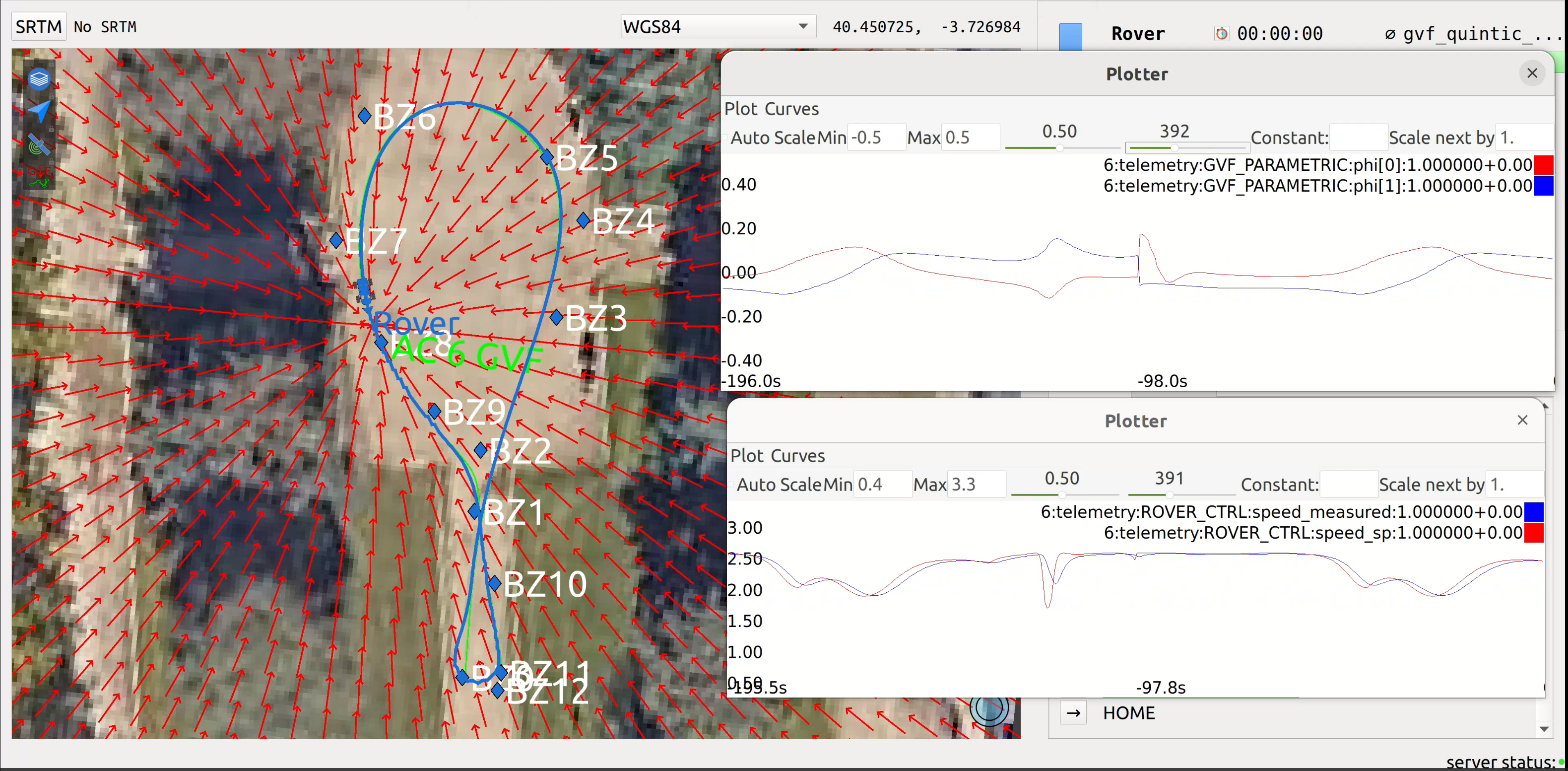}
    \caption{Screenshot from the Paparazzi's GCS in simulation. Upper right graphic shows the distance errors in meters $(\phi_1,\phi_2)$ to the desired path $\mathcal P$ (as phi[0] and phi[1]) respectively) versus time. The bottom right graphic shows the time evolution of $v_{ref}$ and $v$ as red and blue lines in m/s. Constants $k_1=k_2=0.5\text{ m},k_\theta=3$ for the guidance controller and $v_{min} = 1.7 \text{ m/s}, v_{max} = 2.7 \text{ m/s}, c_\kappa = 10\text{ m}^{2}$ for the speed controller were used.}
\label{fig:pprz_simu_new}
\end{figure} 
An example of a curve is created and shown in figure \ref{fig:pprz_simu_new}, where the trajectory intersects with itself and has a changing curvature, necessary conditions to test our controllers. After launching the simulation using the model from equation \eqref{eq:model}, it can be seen how the path traced by the Rover converges to the desired parametric curve and errors $\phi_1,\phi_2$ are in the order of centimetres. Moreover, it can be seen how the speed setpoint (speedsp) and Rover's speed (speed measured) change according to the path's curvature.
Note that the algorithm is restarted when reaching the ending point $\beta_{11}$ by resetting the path parameter $w$ to the initial value $w =0$.

\subsection{Real Rover Results}
This subsection is divided into two parts. The first deals with the actual experiments on the guidance algorithm, while the second deals with the speed control problem. It is also important to note that the following experiments can be reproduced by compiling and downloading the code to the vehicle (provided they have the same or compatible hardware) using the repositories presented above.
%% Begining of images
%\begin{figure}[t]
%    \centering
%    \includegraphics[width=0.6\linewidth]{Figures/FiguresFromLastExperiment/PprzGCS_real_field_last.png}
%    \caption{Screenshot from the Paparazzi's GCS in a real experiment. The representations and meanings are the same as in the simulation. The same constants were used as in the simulation.}
%    \label{fig:pprz_real_gcsField}
%\end{figure}

\begin{figure}
    \begin{minipage}[c]{0.47\linewidth}
    \includegraphics[width=\linewidth]{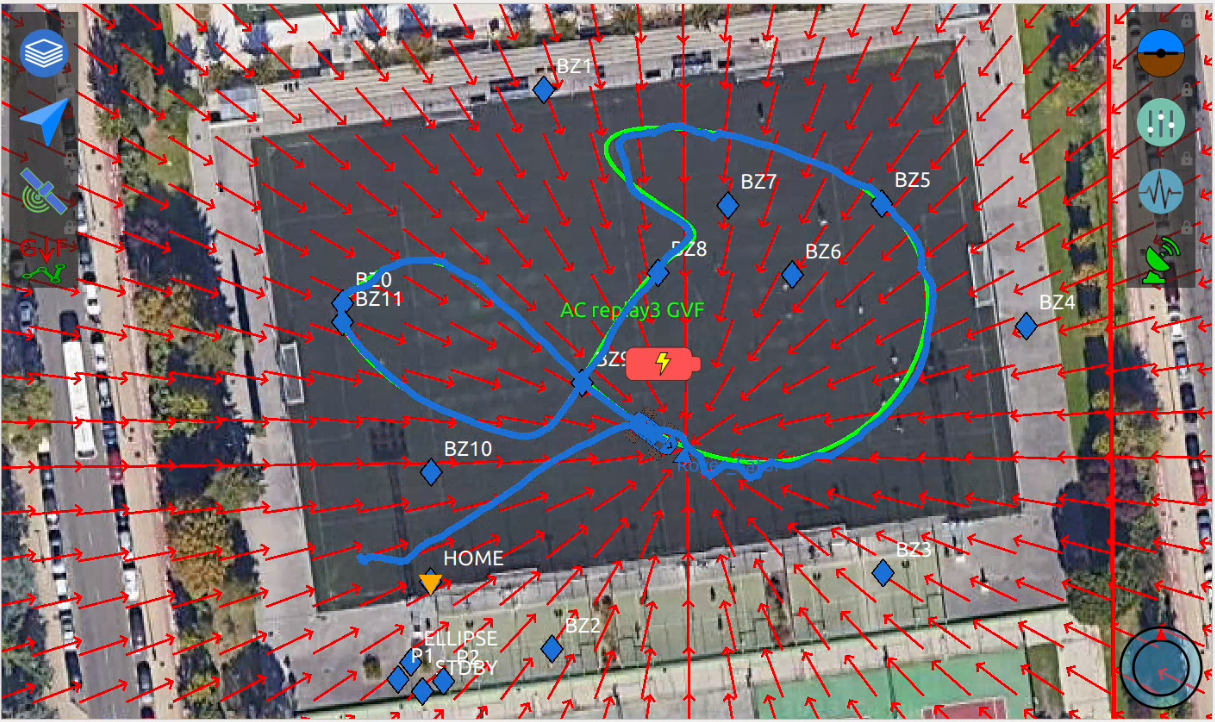}
    \caption{Screenshot from the Paparazzi's GCS in the first real experiment. The representations and meanings are the same as in the simulation. The same constants were used as in the simulation.     \label{fig:pprz_real_gcsField}}
    \end{minipage}
    \hfill
    \begin{minipage}[c]{0.47\linewidth}
    \includegraphics[width=\linewidth]{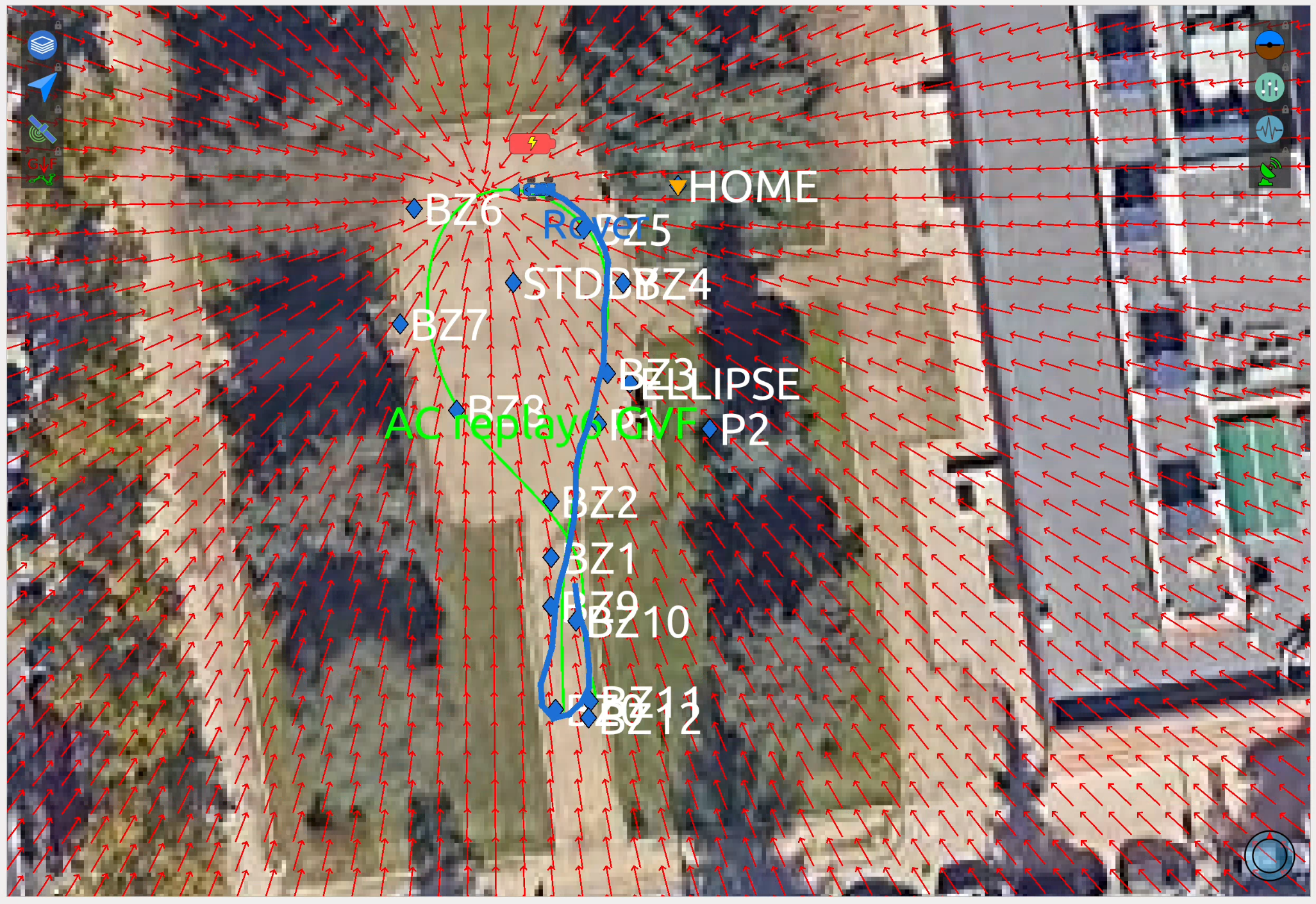}
    \caption{Screenshot from the Paparazzi's GCS in the second real experiment. The same constants
    were used as in the simulation. \label{fig:pprz_real_gcsField_2}}
    \end{minipage}%
\end{figure}

Two self-intersecting curves are chosen for
the SF-GVF and heading controller experimental tests. The first one is an experimental setup with fair environmental conditions, i.e., a clear sky for the GPS receiver and an obstacle-free football field. The second one represents a worse-case scenario, where the area in which the Rover can move is tighter, and the GPS is not providing the best possible resolution. Now, the
speed setpoint is the same for both experiments and is implemented as in equation \eqref{eq:setpoint} using: $v_{min} = 1.4 \text{ m/s}, v_{max} =2.4 \text{ m/s}, c_\kappa = 15 \text{m}^2$. 
Using the Paparazzi replay utility, real data results from both experiments are reproduced in figure \ref{fig:pprz_real_gcsField} for the first case and figure \ref{fig:pprz_real_gcsField_2} for the second one.

\begin{figure}
    \centering
    \includegraphics[width=\linewidth]{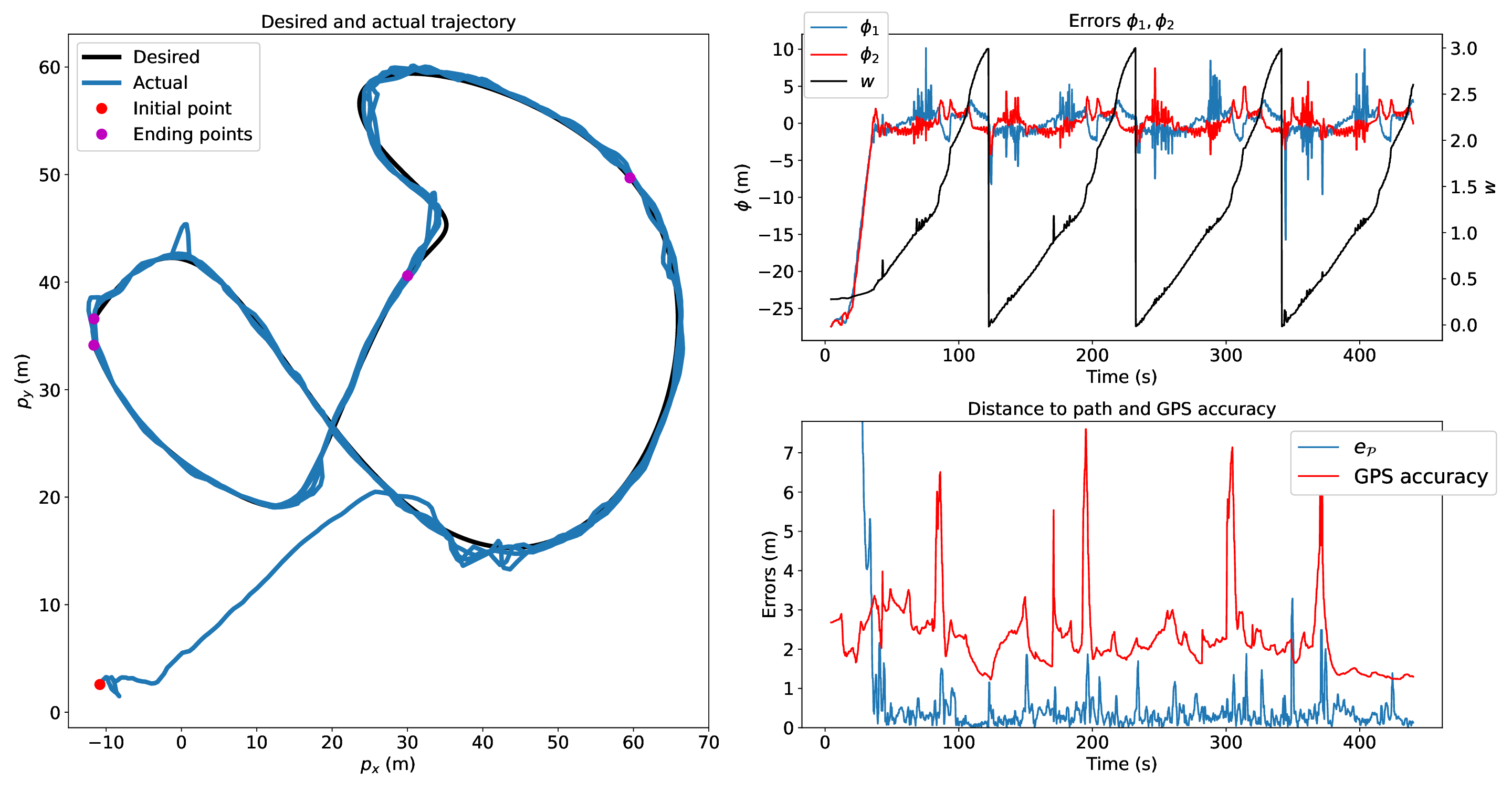}
    \caption{Experimental data obtained from the logs for a complete set of trajectories traced
    by the Rover. The left subfigure shows the actual traced trajectory in blue,
    the desired one in black, the initial point in red, and the curve points
    $\beta_0,\beta_5,\beta_8,\beta_{11}$ in magenta. The upper right subfigure
    shows the errors $\phi_1,\phi_2$ in the left y-axis and the parameter $w$ in
    the right y-axis. The bottom right subfigure shows the distance to the path
    $e_{\mathcal P} = \inf\{||p-p_d|| \mid p_d \in \mathcal P\}$ and the
    accuracy of the GPS \textit{position}. }
    \label{fig:GVF_Football}
\end{figure}

\begin{figure}
    \begin{minipage}[c]{0.45\linewidth}
    \includegraphics[width=\linewidth]{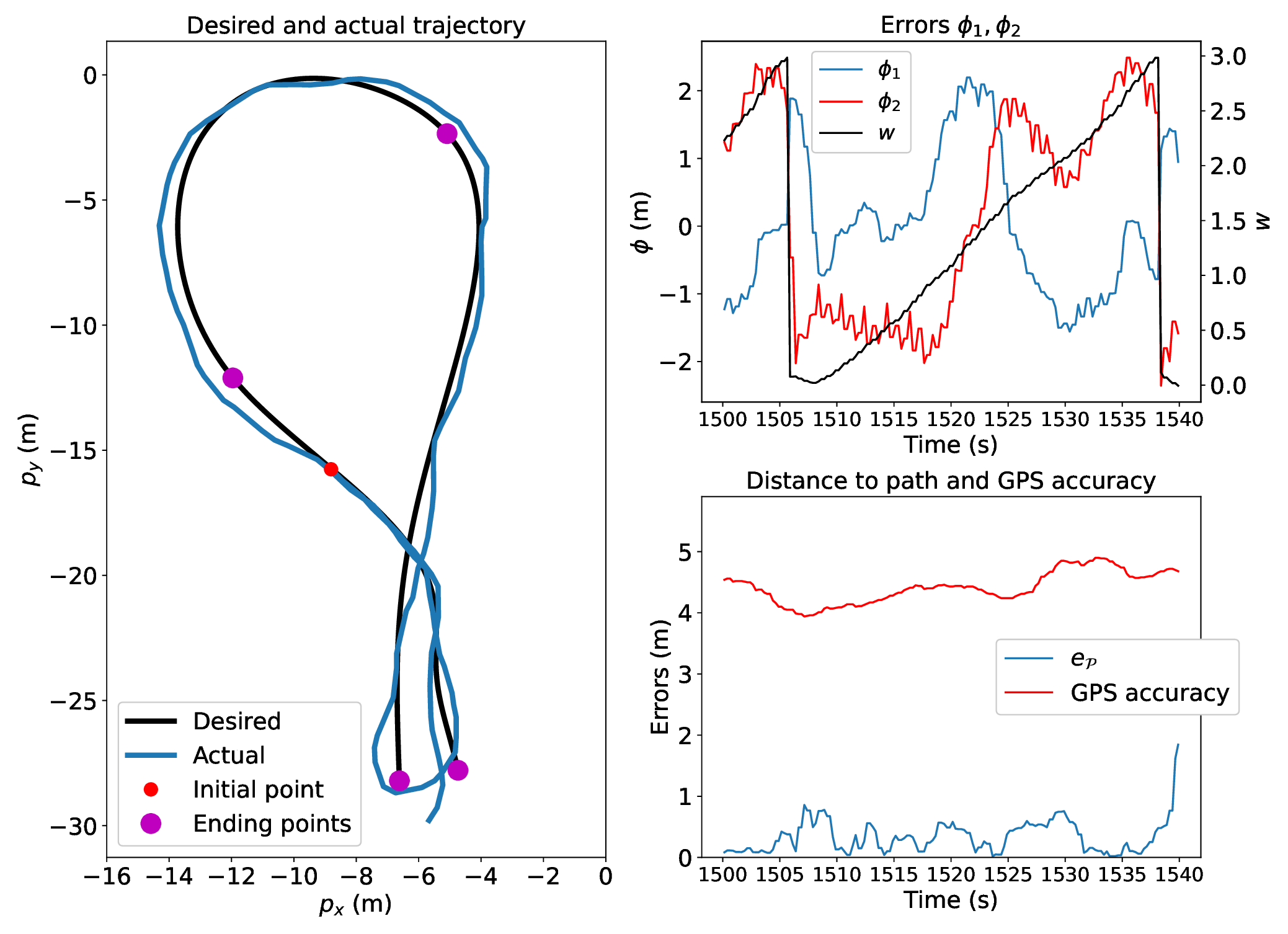}
    \caption{Experimental data obtained from the logs. The left subfigure shows
    a set of actual trajectories and the desired one, the initial point in red,
    and the curve points $\beta_0,\beta_5,\beta_8,\beta_{11}$ in magenta, the
    upper right subfigure shows in the left y-axis the errors $\phi_1, \phi_2$, and in
    the right y-axis the parameter $w$. The bottom right subfigure shows the distance to the
    path $e_{\mathcal P} = \inf\{\lvert p-p_d\rvert \mid p_d \in \mathcal P\}$}, and the GPS \textit{position} accuracy. \label{fig:phierrors}
    \end{minipage}
    \hfill
    \begin{minipage}[c]{0.45\linewidth}
    \includegraphics[width=\linewidth]{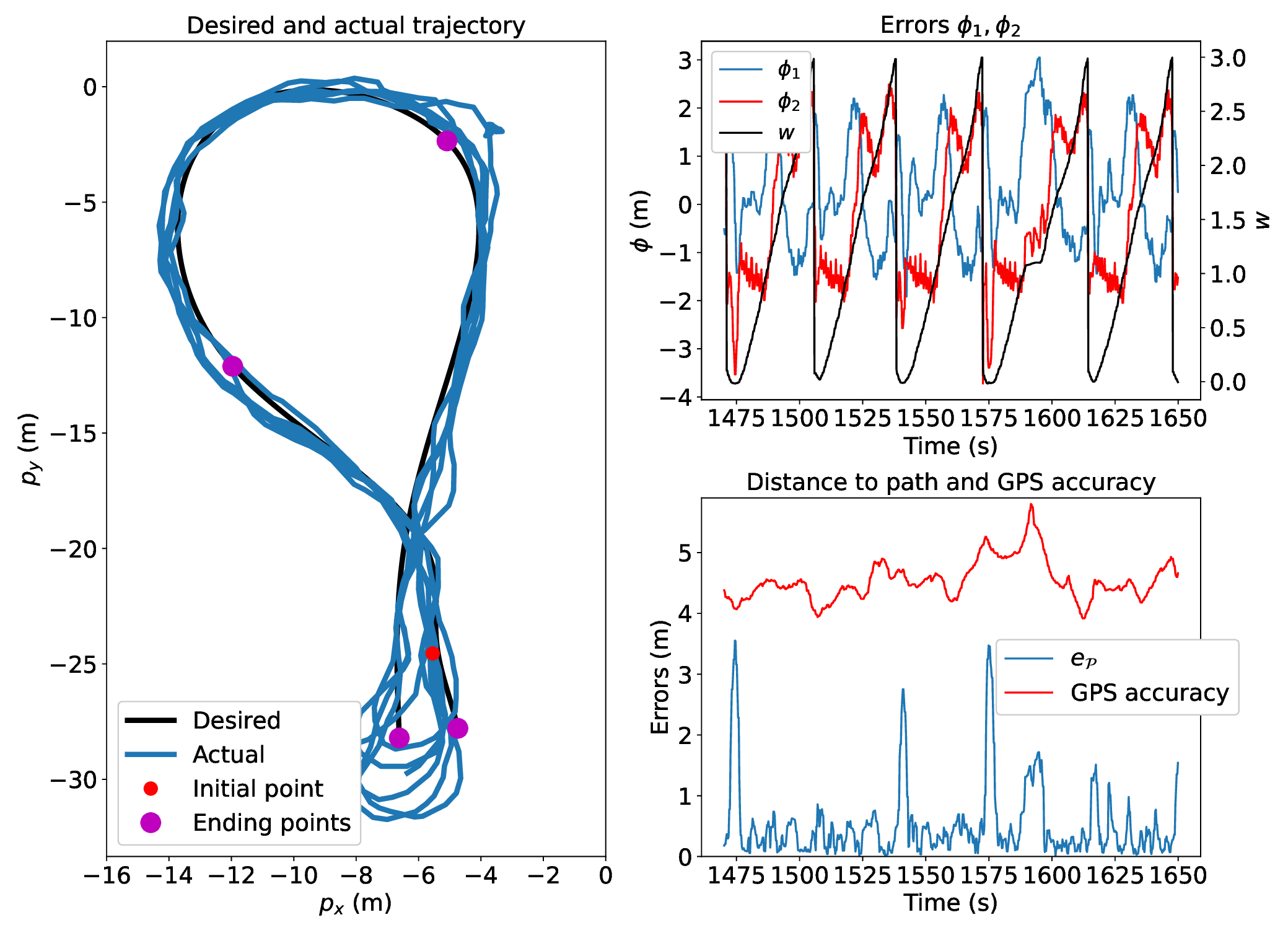}
    \caption{Experimental data obtained from the logs of the same experiment,
    but or a complete set of trajectories. The left subfigure shows a set of
    actual trajectories and the desired one, the initial point in red, and the
    curve points $\beta_0,\beta_5,\beta_8,\beta_{11}$ in magenta, the
    upper right subfigure shows in the left y-axis the errors $\phi_1, \phi_2$, and in
    the right y-axis the parameter $w$. The bottom right subfigure shows the distance to the
    path $e_{\mathcal P} = \inf\{\lvert p-p_d\rvert \mid p_d \in \mathcal P\}$}, and the GPS \textit{position} accuracy. \label{fig:phierrorsv2}
    \end{minipage}%
\end{figure}

Paparazzi replay utility tool allows us to select time intervals of interest to extract data for their analysis. Besides, Paparazzi logs the vehicle's status with the Zigbee telemetry radio. Hence, data from the experiments can be extracted for its posterior analysis. Thus, data from the logs were extracted for both curves to determine the magnitude of the errors. This data are displayed for a complete set of trajectories traced by the Rover for the first experiment curve in figure \ref{fig:GVF_Football}, and for the second one in figures \ref{fig:phierrors} and \ref{fig:phierrorsv2}. 

Figure \ref{fig:phierrors} shows a particular trajectory of the second experiment, while figure \ref{fig:phierrorsv2} shows a set of trajectories made during the same experiment.

%To determine the magnitude of the errors, data from the logs are extracted and displayed in figures \ref{fig:phi errors}, \ref{fig:phi errors_v2}. Figure \ref{fig:phi errors} shows a particular trajectory of an experiment, while figure \ref{fig:phi errors_v2} shows a set of trajectories made during the same experiment.
%choose

For the first experiment,
in figure \ref{fig:GVF_Football}, the left graphic represents the actual trajectory as a blue line and the desired trajectory as a black line where the origin of coordinates is centred on the HOME point shown in figure \ref{fig:pprz_real_gcsField}. Thus, the Rover position $(p_x,p_y)$ is measured relative to the HOME origin attached to the earth and fixed throughout the whole experiment. Home location is determined by the first geographical position the GPS measures.

The Rover begins 25 meters away from the intended path (as shown on the left side graphic of figure \ref{fig:GVF_Football}). The algorithm then guides the Rover towards the path until it reaches it. Once the final segment of the path is reached, the trajectory begins again, the algorithm restarts and $w$ is reset to 0. This is indicated by the Rover converging back to the starting point of the path. The upper right graphic shows the evolution of the errors $\phi_1$ and $\phi_2$ (on the left y-axis) and the parameter $w$ (on the right y-axis) over time. The bottom left graphic displays the error to the path, $e_{\mathcal P} = \inf\{||p-p_d|| \mid p_d \in \mathcal P\}$, as well as the GPS position accuracy.

Once the Rover has converged to the trajectory, the distance to the path $e_{\mathcal P}$ is close to zero and well below the GPS position accuracy throughout the experiment. Note, however, that the errors $\phi_1,\phi_2$, and thus the previously mentioned error $||e(\xi)||$ with $e(\xi) = (\phi_1(\xi), \phi_2(\xi))$, are bigger, and in certain time instants they can reach errors of  10 meters. This behaviour is due to oscillations in the parameter $w$, whose dynamics were specified in equation \eqref{eq:w_def}.

As can be seen in the upper right graphic, the time instants in which $w$ oscillates match those times when the errors $\phi_1,\phi_2$ oscillate and separate from zero. Recall that $\phi_1 = p_x-f_1(w)$ and $\phi_1 = p_y-f_2(w)$ and, for a three segment Bézier spline, $w \in [0,3]$. Thus, even if the Rover has converged to the trajectory, a fast oscillation of $w$ creates fast oscillations of $f_1(w)$ and $f_2(w)$, which in turn creates fast oscillations of $\phi_1$ and $\phi_2$. Moreover, since $w\in[0,3]$ and the dimensions of the Bézier spline are in the order of tens of meters, a fast and small variation of $w$ can create a big variation in $f(w)$ and a big
and fast transient in $\phi_1$ and $\phi_2$. Nevertheless, the distance to the path, $e_{\mathcal P}$, is not directly affected by $w$, and from the left graphic and the bottom right graphic, it can be seen that it converges towards zero and remains below the GPS accuracy.

%In figure \ref{fig:phi errors}, the left graphic represents the actual trajectory as a blue line and the desired trajectory as a black line where the origin of coordinates is centred on the HOME waypoint shown in figure \ref{fig:pprz_real_gcsField}. Thus, the Rover position $(p_x,p_y)$ is measured relative to the HOME origin attached to the earth, fixed throughout the whole experiment, and its location is determined by the first geographical position the GPS measures. It can be seen that the Rover starts two meters away from the desired path, and the algorithm converges to it. Moreover, at the endpoint of the final segment of the spline, the trajectory starts again; that is, $w$ is reset to 0. For that reason, the algorithm restarts, as indicated by the Rover converging to the initial points of the spline. In addition, the distance errors $\phi_1, \phi_2$ are shown in the upper-right figure, while the error function $||e(\xi)||$  and the GPS position accuracy are shown in the bottom left figure. Considering that the error function $||e(\xi)||$ is always below the GPS position accuracy once the Rover has converged to the trajectory, the path is followed correctly. This also means that the error function $||e(\xi)||$ has an offset due to the GPS measurements. 
%
% Adding experiments from football field

% Maybe in one single image ??

For the second experiment, figure \ref{fig:phierrorsv2} represents the same information as figure \ref{fig:phierrors} but for a set of trajectories. Note that, compared to figure \ref{fig:GVF_Football}, the area in which the Rover must remain is tighter, as can be seen from the dimensions of the trajectory. Moreover, the mean GPS accuracy is close to the one in figure \ref{fig:GVF_Football}, allowing for a similar error if the GPS accuracy is considered as the bound of the $d$ term of equation \eqref{eq:ErrorWithdistirbance}. As explained above, the algorithm restarts when reaching the end point of the spline, which means that the trajectories shown in figure \ref{fig:phierrorsv2} are done sequentially, that is, in the same experiment and one after another. 

In this more extensive set of experiments, similar to the previous one, the error is bounded from above throughout the entire experiment by the GPS position accuracy; as it is reflected in the comparison between the desired and actual trajectory. Moreover, at the starting points, it can be seen that a transient loop occurs. This is due to the asymptotic stability of the closed loop system; when the curve is restarted, it shall converge to the trajectory again.
Moreover, comparing with figure \ref{fig:GVF_Football}, this experiment did not show any
fast oscillations of the parameter $w$, and in turn no fast oscillations of the errors $\phi_1$ and 
$\phi_2$.

For
completeness, let us use the results of equation \ref{eq:UpperboundError}, where we arrived at the result
that $||e(\xi)|| \leq \sup_{t\in[0,\infty)}\frac{||d(t)||}{\sqrt{\lambda_{min}}}$. The $Q$ matrix can be 
expressed as
\begin{equation*}
    Q  =\begin{bmatrix}
        q_{11} & q_{12} \\
        q_{12} & q_{22}
    \end{bmatrix} = \begin{bmatrix}
        (f_1'(w)^2 + 1)k_1^2 & f_1'(w)f_2'(w)k_1k_2 \\
        f_1'(w)f_2'(w)k_1k_2 & (f_2'(w)^2+1)k_2^2
    \end{bmatrix} \succ 0.
\end{equation*}
The eigenvalues can be computed to obtain $\lambda = \frac{{q_{11}+q_{22}} \pm \sqrt{(q_{11}+q_{22})^2-4(q_{11}q_{22}-q_{12}^2)}}{2}.$ Moreover, note that in both experiments, we have used the same values of $k_1$ and $k_2$, i.e., $k_1=k_2=0.5$, so plugin in this condition on the eigenvalues and
the values of the $Q$ matrix, we arrive at
\begin{equation*}
    \lambda = \frac{k_1^2(f_1'(w)^2+f_2'(w)^2+2) \pm \sqrt{k_1^4(f_1'(w)^2+f_2'(w)^2)^2}}{2} = \frac{k_1^2(f_1'(w)^2+f_2'(w)^2+2)\pm k_1^2(f_1'(w)^2+f_2'(w)^2)}{2}.
\end{equation*}
Thus, it can be easily seen that $\lambda_{min} = k_1^2, \lambda_{max} = k_1^2(||f'(w)||^2+1)$. So the error can now be bounded, $||e(\xi)|| \leq \sup_{t\in[0,\infty)}\frac{||d(t)||}{\sqrt{\lambda_{min}}} = \sup_{t\in[0,\infty)}\frac{||d(t)||}{\sqrt{k_1^2}}$, so $||e(\xi)|| \leq 2\sup_{t\in[0,\infty)}||d(t)||$.
For the first experiment $\sup ||d(t)|| = 7.6$, while for the second $\sup ||d(t)|| = 5.3$,
so for the first experiment $||e(\xi)|| \leq 15.2 \text{ m}$, while for the second
$||e(\xi)|| \leq 10.6 \text{ m}$. This holds true in both experiments when computing $||e(\xi)|| = \sqrt{\phi_1(\xi)^2 +\phi_2(\xi)^2}$, which connect theoretical
and experimental results. Note that in the first experiment, the supremum of the disturbance is larger (i.e., there are instants in which the GPS uncertainty is larger), while the 
average GPS position error is lower than in the second experiment, as can be seen by comparing figure \ref{fig:GVF_Football} with figures \ref{fig:phierrors} or \ref{fig:phierrorsv2}.

Before commenting on the results of the speed controller, it is worth noting that in the Paparazzi platform, the throttle control actions $u_v$ are bounded by $u_v\in [-9600,9600]$, and those values are converted to PWM signals as explained above. 

\begin{figure}[t]
    \centering
    \includegraphics[width=0.8\linewidth]{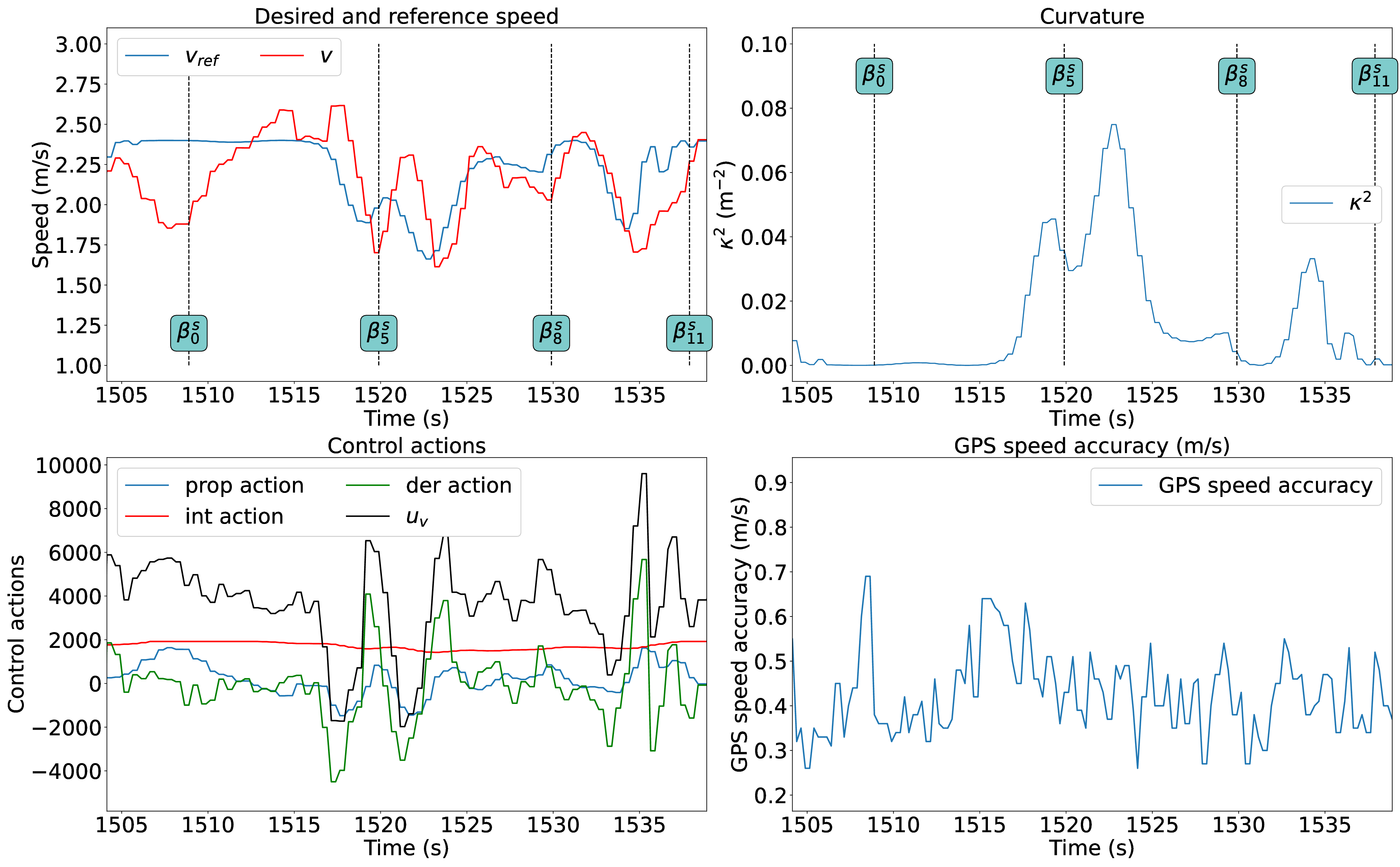}
    \caption{Rover's speed data obtained from the same experiment as in figure \ref{fig:phierrors}. The top left graphic shows the desired speed setpoint as a blue line, the Rover's measured speed as a red line and as vertical black dotted lines the timestamps in which the Rover passes through the corresponding Bézier points $\beta^s_i$. The top right graphic represents the curvature as a function of time $\kappa (w(t))$. The bottom left graphic represents the proportional, integral and derivative control actions and the total speed control action $u_v$. The bottom right graphic represents the \textit{speed} accuracy given by the GPS. Constant used for this experiment: $v_{min} = 1.4 \text{ m/s}, v_{max} =2.4 \text{ m/s}, c_\kappa = 15 \text{m}^2$, $k_f = 1000, k_p = 3000, k_i = 300$ and $k_d = 2000$.}
    \label{fig:speeddata}
\end{figure}

Speed controller data from the second set of experiments is shown in figure \ref{fig:speeddata}. The top left graphic shows the speed setpoint $v_{ref}$ and the GPS speed $v$ filtered with a moving average filter (a low pass filter) of $M = 200$ samples. Vertical black dotted lines represent the timestamps in which the Rover passes through the corresponding Bézier points $\beta^s_i$. The top right graphic depicts the curvature of the Bézier spline. As it can be seen from both top graphics, the speed setpoint $v_{ref}$ changes accordingly to the computed curvature $\kappa$, reducing the setpoint when $\kappa^2$ increases and vice versa. The bottom left graphic shows the speed control signal $u_v$ and its PID components, and the bottom right graphic represents the \textit{speed} accuracy given by the GPS. 

The time delay that can be seen in the top left graphic, where the actual speed $v$ is behind the setpoint, is due to the speed controller settling time and the filter's effect. However, the delay in some way is not an undesired property. At any given time t, the vector field points towards a point of the Bézier curve ahead of the Rover, as shown in figures \ref{fig:pprz_real_gcsField} and \ref{fig:pprz_real_gcsField_2}. The speed setpoint changes based on the curvature of the function $f(w)$; however, assuming the Rover has already reached the trajectory at time $t$, it will be at a point $f(w - w^*)$ where $w-w^* < w$. This indicates that the Rover will reach the point $f(w)$, for which the setpoint was originally calculated after some delay. Thus, the trade-off between noise and time delay can be adjusted with the speed controller constants and the number of samples $M$. In particular, the constants we used are shown in table \ref{tab:BzierPoints}.
Finally, for completeness, tables \ref{tab:BzierPointsFirst} and
\ref{tab:BzierPoints} show the Bézier points and constant values used
in the first and second set of experiments, respectively. Note
that except for the first segment (segment $i = 0$), the points $\beta_1^i,
\beta_2^i$ for $i \in \{1,2\}$ are computed using the $C^2$ continuity
conditions shown in section \ref{sec:PathPlanning}. Thus, the user only needs to
specify the points in figure \ref{fig:pprz_real_gcsField}.

\begin{table}[!h]
\centering
\resizebox{\textwidth}{!}{%

\begin{tabular}{|cccc||cc|}
\hline
\multicolumn{4}{|c||}{Bézier points used for the first experiment} &
  \multicolumn{2}{c|}{Controller constants} \\ \hline
\multicolumn{1}{|c|}{\begin{tabular}[c]{@{}c@{}}Bézier points\\ (w.r.t HOME)\end{tabular}} &
  \multicolumn{1}{c|}{\begin{tabular}[c]{@{}c@{}}Segment 0\\ (m)\end{tabular}} &
  \multicolumn{1}{c|}{\begin{tabular}[c]{@{}c@{}}Segment 1\\ (m)\end{tabular}} &
  \begin{tabular}[c|]{@{}c@{}}Segment 2\\ (m)\end{tabular} &
  \multicolumn{1}{c|}{Constant used} &
  Values \\ \hline
\multicolumn{1}{|c|}{$\beta_0^i$} &       % Beta_0 for segment i
  \multicolumn{1}{c|}{$(-11.62,36.58)$} & % Segment 0
  \multicolumn{1}{c|}{$(59.53, 49.69)$} & % Segment 1
  $(30.12, 40.59)$ &                    % Segment 2
  \multicolumn{1}{c|}{$k_1,k_2$} &
  0.5 \\ \hline
\multicolumn{1}{|c|}{$\beta_1^i$} &        % Beta_1 for segment i
  \multicolumn{1}{c|}{$(14.93, 64.67)$} & % Segment 0
  \multicolumn{1}{c|}{$(40.45,65.78)$} &   % Segment 1
  $(20.78, 31.71)$ &                      % Segment 2
  \multicolumn{1}{c|}{$k_\theta$} &
  1 \\ \hline
\multicolumn{1}{|c|}{$\beta_2^i$} &
  \multicolumn{1}{c|}{$(16.02, -8.84)$} &
  \multicolumn{1}{c|}{$(-16.64,65.54)$} &
  $(10.78,4.84)$ &
  \multicolumn{1}{c|}{$k_f$} &
  1000 \\ \hline
\multicolumn{1}{|c|}{$\beta_3^i$} &
  \multicolumn{1}{c|}{$(59.72,1.15)$} &
  \multicolumn{1}{c|}{$(47.74,40.36)$} &
  $(20.00, 26.13)$ &
  \multicolumn{1}{c|}{$(k_p,k_i,k_d)$} &
  $(3000,300,2000)$ \\ \hline
\multicolumn{1}{|c|}{$\beta_4^i$} &
  \multicolumn{1}{c|}{$(78.63,33.59)$} &
  \multicolumn{1}{c|}{$(39.26,49.47)$} &
  $(0.07, 14.38)$ &
  \multicolumn{1}{c|}{$c_\kappa$} &
  15 \\ \hline
\multicolumn{1}{|c|}{$\beta_5^i$} &
  \multicolumn{1}{c|}{$(59.54,49.69)$} &
  \multicolumn{1}{c|}{$(30.02,40.59)$} &
  $(-11.63,34.13)$ &
  \multicolumn{1}{c|}{$M$} &
  200 \\ \hline
\end{tabular}%
}

\caption{Bézier control points measured from HOME position and constants used for the first experiment.}
\label{tab:BzierPointsFirst}
\end{table}

\begin{table}[!h]
\centering
\resizebox{\textwidth}{!}{%
\begin{tabular}{|cccc||cc|}
\hline
\multicolumn{4}{|c||}{Bézier points used for the second experiment} &
  \multicolumn{2}{c|}{Controller constants} \\ \hline
\multicolumn{1}{|c|}{\begin{tabular}[c]{@{}c@{}}Bézier points\\ (w.r.t HOME)\end{tabular}} &
  \multicolumn{1}{c|}{\begin{tabular}[c]{@{}c@{}}Segment 0\\ (m)\end{tabular}} &
  \multicolumn{1}{c|}{\begin{tabular}[c]{@{}c@{}}Segment 1\\ (m)\end{tabular}} &
  \begin{tabular}[c|]{@{}c@{}}Segment 2\\ (m)\end{tabular} &
  \multicolumn{1}{c|}{Constant used} &
  Values \\ \hline
\multicolumn{1}{|c|}{$\beta_0^i$} &
  \multicolumn{1}{c|}{$(-6.61,-28.20)$} &
  \multicolumn{1}{c|}{$(-5.08, -2.34)$} &
  $(-11.95, -12.10)$ &
  \multicolumn{1}{c|}{$k_1,k_2$} &
  0.5 \\ \hline
\multicolumn{1}{|c|}{$\beta_1^i$} &
  \multicolumn{1}{c|}{$(-6.85, -20.01)$} &
  \multicolumn{1}{c|}{$(-7.20, 0.55)$} &
  $(-8.89, -16.75)$ &
  \multicolumn{1}{c|}{$k_\theta$} &
  1 \\ \hline
\multicolumn{1}{|c|}{$\beta_2^i$} &
  \multicolumn{1}{c|}{$(-6.86, -17.00)$} &
  \multicolumn{1}{c|}{$(-12.26, -1.49)$} &
  $(-2.00, -19.81)$ &
  \multicolumn{1}{c|}{$k_f$} &
  1000 \\ \hline
\multicolumn{1}{|c|}{$\beta_3^i$} &
  \multicolumn{1}{c|}{$(-3.80, -10.11)$} &
  \multicolumn{1}{c|}{$(-14.25, -1.25)$} &
  $(-6.86, -22.65)$ &
  \multicolumn{1}{c|}{$(k_p,k_i,k_d)$} &
  $(3000,300,2000)$ \\ \hline
\multicolumn{1}{|c|}{$\beta_4^i$} &
  \multicolumn{1}{c|}{$(-2.97, -5.25)$} &
  \multicolumn{1}{c|}{$(-15.01, -7.46)$} &
  $(-5.51, -23.45)$ &
  \multicolumn{1}{c|}{$c_\kappa$} &
  15 \\ \hline
\multicolumn{1}{|c|}{$\beta_5^i$} &
  \multicolumn{1}{c|}{$(-5.08, -2.34)$} &
  \multicolumn{1}{c|}{$(-11.95, -12.10)$} &
  $(-4.73, -27.78)$ &
  \multicolumn{1}{c|}{$M$} &
  200 \\ \hline
\end{tabular}%
}
\caption{Bézier control points measured from HOME position and constants used for the second experiment.}
\label{tab:BzierPoints}
\end{table}

%printbibliography hasn't the style 
%\printbibliography
\section{Conclusions}\label{sec:conclusions}
In this paper, a guidance system based on GVFs is designed for path-following in AMR vehicles, and a curvature-dependent speed controller is used for faster convergence of the guidance algorithm. The definition of the desired path as parametric curves allows for a straightforward computation of the curve's curvature for the speed setpoint. In addition to presenting theoretical work involving these two controllers, simulations of the complete system are made using the Paparazzi UAV platform. Finally, we introduce our AMR platform, for which exhaustive experiments are made to show the approach's effectiveness.
The main contributions of this work are:
 \begin{itemize}
    \item Generate a waypoint trajectory using Bézier curves, allowing the human user to easily set the waypoints and control points on the map and change them on the fly if necessary. Bézier curves allow smooth trajectories to be generated within the defined bounding box, formally called a convex hull. 
    \item Once the trajectory is established, GVF is employed to define a vector field that allows us to know, at any point in space, the desired velocity direction of the rover that will allow it to converge to the target path. The vector field can be applied to derive a control law for the convergence property. GVF algorithms show exemplary performance. To the authors' knowledge, no results of this Singularity Free GVF (SF-GVF) applied to the Rover path-following problem exist.
    \item Using SF-GVF and the Rover kinematic model, a controller is derived using a Lyapunov function, assuring stability. 
    \item Path following is improved by speed control based on path curvature.
    \item Results of the guidance and speed controllers are presented in simulation and with a real Rover in a natural environment. For this purpose, the Paparazzi environment is used \cite{gati2013open}.
 \end{itemize}

The ease and adaptability of Bézier polynomials make them very attractive to extend their use to other autonomous vehicles. For instance, we are currently testing the same approach to control Autonomous Surface Vehicles for water quality monitoring in inland reservoirs. imulation results are presented in \cite{gonzalezcalvin2023}, using a planner and a guidance and control algorithm to traverse the planned trajectory. The results presented in this paper: generation of trajectories from measurement points using Bezier curves, tracking of these trajectories using SF-GVF and velocity control based on the curvature of the path are being tested on a USV with measurement probes for the detection of cyanobacterial blooms. The results are promising and will be the object of future publications.

Besides, we are also interested in extending our algorithm for obstacle avoidance, using data from proximity sensors to modify the shape of the Beziérs curves on the fly. So, the vehicle will replan its path, maintaining the prescribed waypoints wherever possible.  
\section*{Acknowledgments}
This work is partially supported by IA-GES-BLOOM-CM (Y2020/TCS-6420) of the Synergic Projects program of the Autonomous Community of Madrid and INSERTION (PID2021-27648OB-C33) of the Knowledge Generation program of the Ministry of Science and Innovation. 
\bibliographystyle{apalike}
\bibliography{references}
\end{document}